\documentclass[10pt,twocolumn,letterpaper]{article}

\usepackage[pagenumbers]{cvpr} 

\usepackage[dvipsnames]{xcolor}

\usepackage[T1]{fontenc}
\usepackage[utf8]{inputenc}

\usepackage{amsmath, amsfonts, amssymb}
\usepackage{caption}
\usepackage{subcaption}
\usepackage{multirow}
\usepackage{booktabs}
\renewcommand{\arraystretch}{1.3}

\usepackage{color, colortbl}
\definecolor{Gray}{gray}{0.9}

\usepackage{graphicx}

\definecolor{cvprblue}{rgb}{0.21,0.49,0.74}
\usepackage[pagebackref,breaklinks,colorlinks,citecolor=cvprblue]{hyperref}

\def\paperID{10} 
\def\confName{SAIAD}
\def\confYear{2024}

\title{Criteria for Uncertainty-based Corner Cases Detection in Instance Segmentation}

\author{
    Florian Heidecker$^{1}$, 
    Ahmad El-Khateeb$^{1}$, 
    Maarten Bieshaar$^{2}$, 
    Bernhard Sick$^{1}$\\ 
    \and
    $^{1}$ Intelligent Embedded Systems, University of Kassel\\
    Wilhelmshöher Allee 73, 34121 Kassel, Germany\\
    {\tt\small \{florian.heidecker,akhateeb,bsick\}@uni-kassel.de}
    \and
    $^{2}$ Bosch Center for Artificial Intelligence\\
    31132 Hildesheim, Germany\\
    {\tt\small maarten.bieshaar@de.bosch.com}
}

\begin{document}
\maketitle
\begin{abstract}
    The operating environment of a highly automated vehicle is subject to change, e.g., weather, illumination, or the scenario containing different objects and other participants in which the highly automated vehicle has to navigate its passengers safely. These situations must be considered when developing and validating highly automated driving functions. This already poses a problem for training and evaluating deep learning models because without the costly labeling of thousands of recordings, not knowing whether the data contains relevant, interesting data for further model training, it is a guess under which conditions and situations the model performs poorly. For this purpose, we present corner case criteria based on the predictive uncertainty. With our corner case criteria, we are able to detect uncertainty-based corner cases of an object instance segmentation model without relying on ground truth (GT) data. We evaluated each corner case criterion using the COCO and the NuImages dataset to analyze the potential of our approach. We also provide a corner case decision function that allows us to distinguish each object into True Positive (TP), localization and/or classification corner case, or False Positive (FP). We also present our first results of an iterative training cycle that outperforms the baseline and where the data added to the training dataset is selected based on the corner case decision function.
\end{abstract}


\section{Introduction}
    Various driving assistance systems rely on object detection, such as pedestrian or traffic sign detection. Object detection is also essential for highly automated driving, enabling the vehicle to navigate safely among other road users. Object instance segmentation further provides the instance mask besides the object class and bounding box. The object instance segmentation model must work properly even under constantly changing weather, illumination, scenarios, participants, and other conditions, including corner cases. 
    
    Corner Cases~\cite{Heidecker2021b, Bogdoll2021, Roesch2022} are strongly related to anomalies~\cite{Chandola2009, Pimentel2014, Gruhl2021}, outliers~\cite{Hawkins1980, Gruhl2021}, and novelties~\cite{Chandola2009, Gruhl2021} but also cover samples where the model fails~\cite{Pfeil2022, Gawlikowski2022, Heidecker2021b, Houben2021} and data relevant for model improvement \cite{Roesch2022, Breitenstein2023}. In \cite{Heidecker2021b}, a categorization of corner cases for perception in automated driving is introduced, which includes a bunch of potential corner case sources covering the sensor, data content, temporal aspect, and the ML method. Corner cases of the ML method~\cite{Heidecker2021b} are caused due to a lack of knowledge as the model has never encountered a similar sample before, the applied ML method themselves or adversarial samples~\cite{Goodfellow2015}. However, it is difficult to find the data we are looking for because there are no labels, and we do not know whether they represent corner cases for the model and thus provide a value for training to improve the model. 
    
    Corner case detection enables data selection to be guided to identify valuable data and label it more efficiently, offering tremendous cost-saving potential. Besides, there are also other use cases, e.g., active learning~\cite{Kottke2021}, novelty detection~\cite{Gruhl2021, Pimentel2014}, and dataset construction, i.e., creating a training and testing dataset covering all relevant and therefore crucial situations.

    This article presents novel corner case criteria for detecting uncertainty-based corner cases in an object instance segmentation task. These criteria cover the uncertainty in classification, bounding box regression, and instance mask prediction. Our goal is to provide the reader with a clear understanding of how the corner case criteria are made up and function. Furthermore, we evaluate the corner case criteria and thus discuss the limitations and advantages of each corner case criterion. The evaluation is based on two real datasets, COCO~\cite{Lin2015} and NuImages~\cite{Caesar2020}, to show that our corner case criteria are not dataset-specific and can be applied to other datasets. A first approach shows that with our corner case criteria as input feature and a decision function, it is possible to detect corner cases based on the predictive uncertainty and classify predicted objects as True Positive (TP), localization and/or classification corner case, or False Positive (FP) reliably.

    The remainder of this article is structured as follows: Section~\ref{sec:related_work} provides a brief overview of related work. Section~\ref{sec:over} describes the overall approach and provides a basic understanding of using the presented corner case criteria. In Section~\ref{sec:criteria}, we explain each corner case criteria and evaluate their feature importance. Section~\ref{sec:exper} explains the experimental setup. With the corner case criteria, we can classify whether an object is a corner case or not in Section~\ref{sec:decision} and also perform a cycle based model training (cf.~Section~\ref{sec:iter_retrain_model}). Finally, Section~\ref{sec:conclusion} concludes the article’s key message and provides possible future work.

\section{Related Work}\label{sec:related_work}
    Before we continue, we should briefly discuss what uncertainty-based corner cases are and how they are defined. Regarding ML, corner cases are cases caused by an erroneous, malfunction, or incorrect behavior of the trained ML model \cite{Houben2021, Pfeil2022, Gawlikowski2022, Heidecker2021b}. According to the examples given in \cite{Heidecker2021b}, these can be anomalies, outliers, novelties, distribution shifts, and the ML model itself, which delivers erroneous predictions due to the implemented architecture, methodology, training parameters, etc. Investigating uncertainty is important for evaluating the prediction of the model. Moreover, considering model performance, it is possible to reveal information on individual samples or identify samples in which the model is maximally uncertain --- which are corner cases.

    Various approaches already address the challenge of detecting corner cases, such as Out-of-Distribution detection, e.g., using auxiliary Generative Adversarial Networks~\cite{Nitsch2021}, uncertainty-based Out-of-Distribution~\cite{Sedlmeier2019} with Monte-Carlo (MC) Dropout~\cite{Gal2016} or techniques for entropy maximization~\cite{Chan2021}. Lis et al.~\cite{Lis2019}, and Xia et al.~\cite{Xia2020} use image resynthesis to detect pixel anomalies in image data. Bolte et al.~\cite{Bolte2019} also detect corner cases by fusing information from semantic segmentation and object predictions. In ~\cite{Ouyang2021a}, Ouyang et al. present an approach for corner case detection for a classification task using modified distance-based surprise adequacy. Pfeil et al.~\cite{Pfeil2022} use ensembles to detect corner cases in trajectory data, and in \cite{Stocco2020}, a variational autoencoder is used.

    However, we take a different approach by modeling the uncertainty with MC-Dropout~\cite{Gal2016} comparable to \cite{Heidecker2023, Morrison2019, Heidecker2021a, Riedlinger2022}. This allows us to derive the uncertainty with respect to each detected object from the model results. Subsequently, the criteria presented in this article are used to identify corner cases based on the predictive uncertainty and additionally to classify them into different corner case categories (cf. Section~\ref{sec:criteria}).

\section{Overall Approach}\label{sec:over}
    Our approach aims to identify specific corner cases \cite{Heidecker2021b} based on criteria derived from the uncertainty.
    
    \begin{figure}[ht]
        \vspace{5pt}
        \centering
        \includegraphics[width=\linewidth,trim={0 0 0 0},clip]{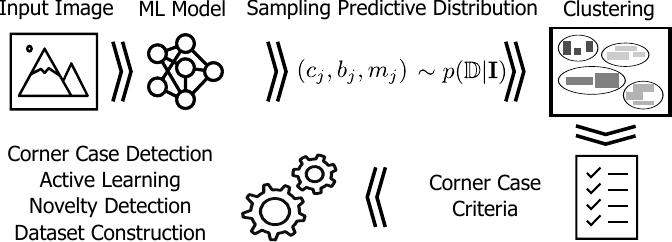}
        \caption{Overall approach to derive and apply uncertainty-based corner case criteria.}\label{fig:overview}
        \vspace{-13pt}
    \end{figure}
    \subsection{Uncertainty and Probabilistic Object Detection}
        We consider a probabilistic definition of object detectors~\cite{Feng2021} and instance segmentation networks~\cite{Hall2020}. Probabilistic object detection extends conventional object detection and instance segmentation by quantifying uncertainty. Let $\mathbb{D} = \{d_{1}, \dots, d_{M}\}$, $M \in \mathbb{N}^+$ denote object detections for a given input image $\mathbf{I}$ with width $W$ and height $H$. A detection $d_j$ incorporating instance segmentation is a tuple $(c_j, b_j, m_j)$\footnote{For object detection, it is a tuple with class score and bounding box.} consisting of information about the object class score $c=(c_{1}\dots c_{k})$ of $k \in \mathbb{N}^+$ classes, bounding box $b = (x_1, y_1, x_2, y_2) \in \mathbb{R}^4$, and binary instance mask $m \in \{0, 1\}^{W \times H}$. Our approach relies on the modeling of the predictive distribution $p(\mathbb{D} | \mathbf{I})$~\cite{Bishop2006}. This predictive distribution includes both the aleatoric and epistemic model uncertainty~\cite{Huellermeier2021}. However, modeling this distribution is challenging due to the high dimensionality of the input and the target space (e.g., there can be any number of predictions in the image). There are numerous methods for approximating this distribution, including Bayesian methods, variational inference, and sample-based representation of the distribution. 
        
        Our approach relies on the latter (i.e., sampling from $p(\mathbb{D} | \mathbf{I})$), cf.~\cite{Heidecker2023, Morrison2019, Heidecker2021a, Riedlinger2022, Feng2021}. There are numerous methods, including MC-Dropout~\cite{Gal2016}, ensembles~\cite{Lakshminarayanan2017}, which also model epistemic uncertainty, or probFRCNN~\cite{Hall2020}, which only considers the aleatoric uncertainty. Our experiments use an instance segmentation model with MC-Dropout (modified Mask R-CNN model~\cite{Heidecker2023}) and cluster multiple forward pass predictions $R$ (repetitions) of the same original input image. Hence, instead of $M$ detections, we end up with $L \in \mathbb{N}^+$ detections with $L \gg M$. In the following, we denote the set of detections (i.e., samples) as $\tilde{\mathbb{D}} = \{d_{1}, \dots, d_{L}\}$. Subsequently, we cluster the samples from the predictive distribution based on the bounding boxes (cf. \cite{Heidecker2023}). After this step we have $i$-th clusters $D_i = \{\{c^{(1)}_i, \dots, c^{(N_i)}_{i}\},\{b^{(1)}_i, \dots, b^{(N_i)}_i\},\{m^{(1)}_i, \dots, m^{(N_i)}_i\}\}$ where $c^{(1)}_i, b^{(1)}_i$, and $m^{(1)}_i$ denote the first detection associated with the $i$-th cluster. Moreover, let $N_i \in \mathbb{N}^+$ be the size of the $i$-th cluster. Clustering the detections from the predictive distribution is the starting point for our uncertainty-driven corner case criteria. An overall schematic representation of our approach to derive uncertainty-based corner case criteria and possible applications using them is depicted in Fig.~\ref{fig:overview}.

        \begin{figure}[t]
            \vspace{5pt}
            \centering
            \includegraphics[width=0.475\textwidth,trim={0 0 1cm 0},clip]{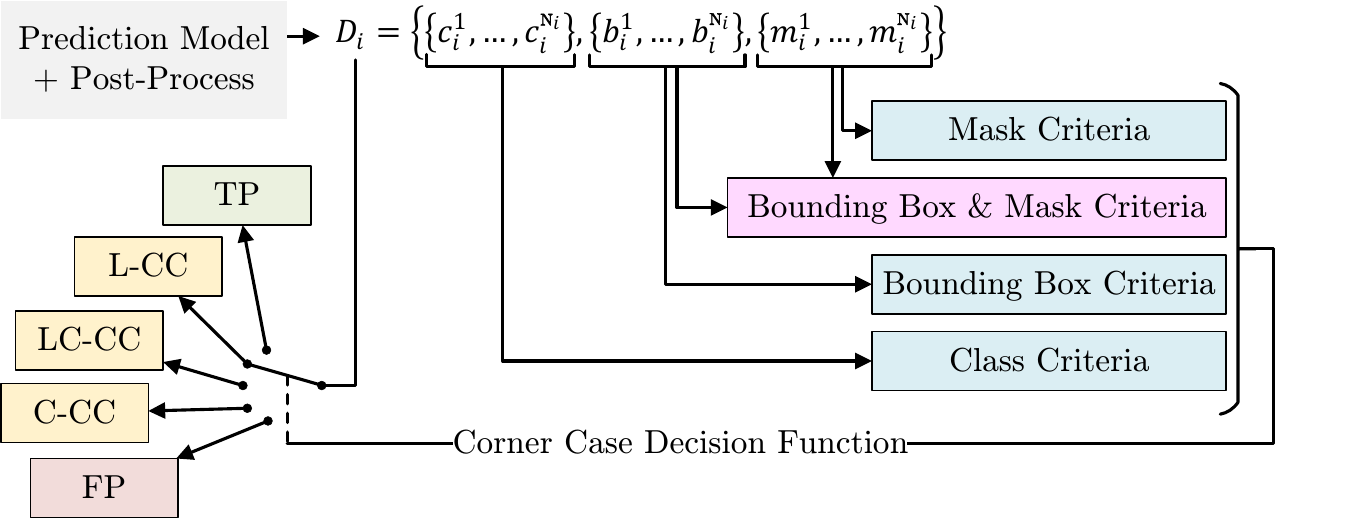}
            \caption{Our approach assumes that the object instance segmentation model plus a possible post-process (gray) provides several predictions, consisting of class score $c$, bounding box $b$, and instance mask $m$, per object detection $D_{i}$. The prediction variance is used to determine the uncertainty in the ML model. The uncertainty analysis is utilized in the class score, bounding box, and instance mask criteria (single knowledge criteria -- blue) or combined in the bounding box \& mask criteria (combined knowledge criteria -- magenta). The values from the criteria are interpreted as features to assign the object detections $D_{i}$ with a decision function to one of the defined corner case categories. The defined categories are True Positive (TP -- green), False Positive (FP -- red), and the corner cases (yellow) for location (L-CC), class (C-CC), and both (LC-CC).}\label{fig:approach}
            \vspace{-15pt}
        \end{figure}

        The key idea behind the uncertainty-based criteria is to use them to detect corner cases. In other words, these criteria should permit us to draw conclusions (in an unsupervised fashion without any labels) about possible deficiencies in the model's performance. In this context, we differentiate between corner case categories introduced later in Section~\ref{subsec:cc_cat}. We derive the criteria based on the clustering and the associated classification scores, bounding boxes, and instance masks. Our corner case criteria can be distinguished into single and combined knowledge criteria; see Fig.~\ref{fig:approach}. As single knowledge criteria in Section~\ref{subsec:class_criteria}, \ref{subsec:box_criteria}, and \ref{subsec:mask_criteria}, we refer to criteria that only use the uncertainty of one feature, e.g., class score criteria, while combined knowledge criteria in Section \ref{subsec:box_mask_criteria} use several sources, e.g., bounding box and mask. Hence, it depends on the application which criterion is useful for the uncertainty analysis. It is also possible to exclude specific criteria if the model provides, for example, no instance mask. 

        Two further steps are necessary to enable the detection of corner cases with the given uncertainty-based criteria. First, the definition of corner case categories, as well as the conditions that must apply so that a detected object may represent a corner case, see Section~\ref{subsec:cc_cat}, and secondly, a decision function that assigns the predicted detection $D_{i}$ to one of the defined corner case based on the criteria, see Section~\ref{sec:decision}.

    \subsection{Corner Cases Categories}\label{subsec:cc_cat}
        To define corner case categories, we need an object-based metric to decide if the detections are corner cases with which the trained model has issues. Typically, we would think of True Positive (TP), False Positive (FP), False Negative (FN), and True Negative as categories, but this separation is much too rough. Hence, we decided to use the $\mathrm{IoU}=\frac{\text{Area of Overlap}}{\text{Area of Union}}=\frac{TP}{TP+FP+FN}$ combined with information if the predicted class is correct or wrong to differentiate the categories. With the metric in place, we defined that a match of ground truth (GT) and detection $D_i$ with an $\mathrm{IoU}>0.5$ and correct class is a good prediction (True Positive (TP)). 
        The threshold of $\mathrm{IoU}=0.5$ is freely selectable and separates good object prediction where no further model improvement is required from the rest. In the future, this threshold might be too low, as a higher IoU value would definitely be preferable. However, objects below an IoU value of $0.5$ are not considered in any mean average precision (mAP) calculations, so we think these samples improve the model the most. We also defined a lower limit of $\mathrm{IoU}=0.1$ because, beyond a certain point, the match between GT and prediction is too low.
        Altogether, we then get the following corner case categories for the class assignment of the object detections:
        \begin{enumerate}
            \item True Positive (TP): Class is correctly predicted, and there is a match between detection $D_i$ and GT with an IoU above $0.5$.
            \item Localization Corner Case (L-CC): Class is correctly predicted, but the match between detection $D_i$ and GT has an IoU between $0.1$ and $0.5$.
            \item Classification Corner Case (C-CC): Class is wrongly predicted, but the match between detection $D_i$ and GT is still above an IoU of $0.5$.
            \item Localization \& Classification Corner Case (LC-CC): Class is wrongly predicted, and the match between detection $D_i$ and GT has an IoU between $0.1$ and $0.5$.
            \item False Positive (FP): All detections $D_i$ with an IoU below $0.1$, which also includes all detections without a GT match.
            \item False Negative (FN): It would be very interesting to know, but there is no way to get this information from the model because there is no prediction to work with.
        \end{enumerate}
        In \cite{Bolya2020, Jia2022}, we find an almost identical categorization with the same metric where they talk about ML model errors in image and video data instead of corner cases as we do. But these model errors, i.e., corner cases, are exactly the problematic samples where the model needs further training samples to improve. On top, the presented categories from \cite{Bolya2020, Jia2022} are used to evaluate the model results better but can also be used for labeling the detections as we do. To then classify the objects using the uncertainty-based corner case criteria as features (cf. Section~\ref{sec:criteria}). Also, the purpose of the categorization differs as \cite{Bolya2020, Jia2022} uses them to compare the ML model evaluation results. Apart from the different naming of the categories, one difference in the categorization is that we do not consider duplicate detections as corner cases —-- we treat them as FP as another detection with a higher IoU exists. 
        \cite{Jia2022} also has another category that deals with a temporal error since video data is considered. This category currently has no meaning because we consider only single images and no sequences. However, this category must be considered for potential corner cases as soon as our methodology is applied to video data and temporal corner cases arising from uncertainty. 

\section{Corner Case Criteria}\label{sec:criteria}
    The corner case criteria are essential to the approach presented in Section~\ref{sec:over}. Each criterion below is calculated based on the knowledge of a single cluster $D_i$, which describes and quantifies the model's uncertainty with a value. Besides, we treat the detected objects separately and do not aggregate them over one image to get one or multiple averaged values for each image.
    
    \subsection{Class Score Criteria}\label{subsec:class_criteria}
        First, we examine the class uncertainty, which greatly influences object detection since it determines which objects are in the image. For the class score, we define four criteria that address the confidence of the class probability. First, we determine the detections' mean class scores $\overline{D}_{c^k}$ and standard deviation $\sigma_{c^k}$ for all classes $k$, see Eq.~\ref{eq:class_mean_std}. Then, the index of $k$ is calculated for the class with the highest ${k_{\mathrm{max}}=\mathrm{argmax}~\overline{D}_{c^k}}$ and second-highest $k_{\mathrm{2nd}}$ class confidence. In summary, we obtain four class score criteria $\overline{D}_{c^{k_{\mathrm{max}}}}$, $\sigma_{c^{k_{\mathrm{max}}}}$, $\overline{D}_{c^{k_{\mathrm{2nd}}}}$, and $\sigma_{c^{k_{\mathrm{2nd}}}}$.
        \begin{equation}\label{eq:class_mean_std}
            \overline{D}_{c^k} = \frac{\sum^{N}_{j=1} c^{k}_{j}}{N},~~\sigma_{c^k}=\sqrt{\frac{\sum^{N}_{j=1} (c^{k}_{j} - \overline{D}_{c^k})^2}{N-1}}
        \end{equation}
        The class probability gives us the same information as any other classification model without uncertainty modeling. Still, in combination with the $\sigma$, we can measure how confident the ML model is in its detections. 

    \subsection{Bounding Box Criteria}\label{subsec:box_criteria}
        \begin{figure}[ht]
            \vspace{5pt}
            \centering
            \includegraphics[width=0.45\textwidth, trim={0, 15pt, 0, 15pt}, clip]{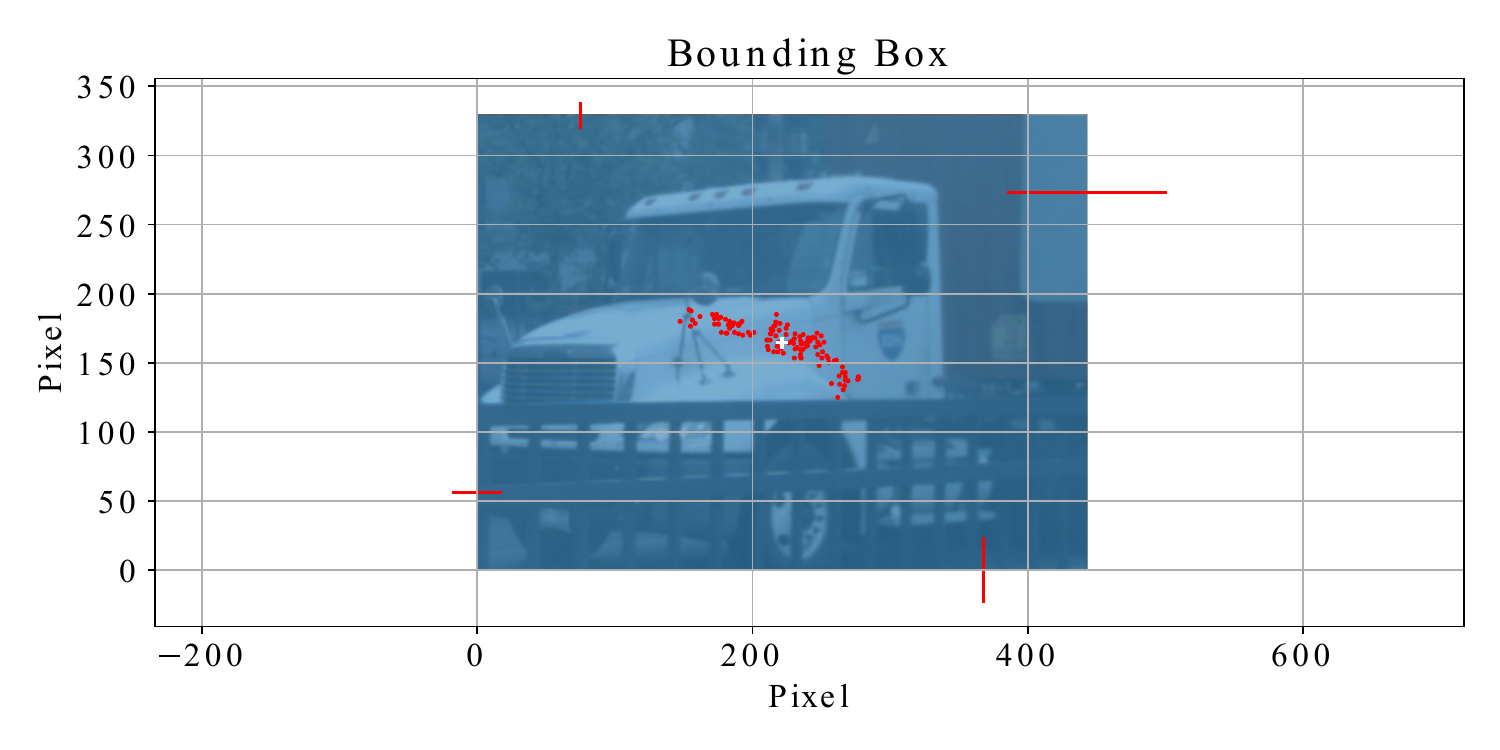}
            \caption{Bounding box uncertainty. The standard deviation $\sigma_{b}$ of each bounding box edge is represented by red lines, and red dots represent the spread of the bounding box center point. The blue box with the truck in the middle represents the mean bounding box.}
            \label{fig:bbox_example}
            \vspace{-10pt}
        \end{figure}
        For the spatial uncertainty of the model's detections, i.e., the bounding box predictions, we primarily consider the standard deviation of the four bounding box edges as a criterion for the bounding box uncertainty. An example of the bounding box uncertainty is depicted in Fig.~\ref{fig:bbox_example}. 
        
        We first calculate the mean $\overline{D}_{b}$ of the bounding box cluster and then the standard deviation $\sigma_{b}$ in Eq.~\ref{eq:box_mean_std}. 
        \begin{equation}\label{eq:box_mean_std}
            \overline{D}_{b} = \frac{\sum^{N}_{j=1} b_{j}}{N},~~\sigma_{b}=\sqrt{\frac{\sum^{N}_{j=1} (b_{j} - \overline{D}_{b})^2}{N-1}}
        \end{equation}
        To see if the standard deviation of the bounding boxes is huge or small compared to its size, we normalize the standard deviation $\sigma_{b}$ based on the mean bounding box size. A low standard deviation $\sigma_{b}$ indicates that the detector predicts the corresponding edge reliably. In contrast, a high standard deviation $\sigma_{b}$ indicates difficulties separating the object from the neighboring object or background. Instead of the format $b=\{x_1,y_1,x_2,y_2\}$ also, $b^*=\{c_x,c_y,w,h\}$ can be used, whereby it is to be considered that both variants behave differently regarding the uncertainty. For example, width $w$ and height $h$ describe only the uncertainty of the bounding box size and no spatial uncertainty in the $x$- or $y$-direction, as all others do. 
        
        We also consider the IoU between the mean bounding box $\overline{D}_{b}$ and all other predictions $b_{j}$ as a further criterion. Therefore, we calculate the mean IoU $\overline{iou}_b$ and the IoU standard deviation $\sigma_{iou_b}$ in Eq.~\ref{eq:box_iou_mean} and \ref{eq:box_iou_std}. 
        \begin{equation}\label{eq:box_iou_mean}
            \overline{iou}_b = \frac{\sum^{N}_{j=1} \mathrm{IoU}(b_{j}, \overline{D}_{b})}{N}
        \end{equation}
        \begin{equation}\label{eq:box_iou_std}
            \sigma_{iou_b}=\sqrt{\frac{\sum^{N}_{j=1} (\mathrm{IoU}(b_{j}, \overline{D}_{b}) - \overline{iou}_b)^2}{N-1}}
        \end{equation}
        The mean value $\overline{iou}_b$ tells us how significant the overlap is between the individual bounding boxes within a cluster. The larger the value, the less uncertain the ML model is about its predictions. The standard deviation of the IoU $\sigma_{iou_b}$, on the other hand, provides information about the spread of the IoUs.

    \subsection{Instance Mask Criteria}\label{subsec:mask_criteria}
        \begin{figure}[ht]
        \vspace{5pt}
            \centering
            \begin{subfigure}[h]{0.45\textwidth}
                \centering
                \includegraphics[width=0.45\textwidth,trim={0 0 60cm 0},clip]{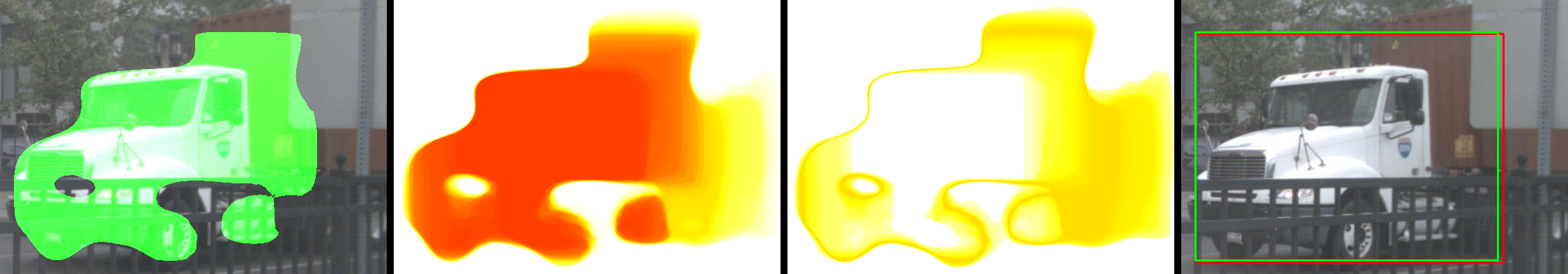}
                \includegraphics[width=0.45\textwidth,trim={60cm 0 0 0},clip]{images/obj_0_heatmap.png}
                \caption{The left shows the binary mask mean $\overline{D}_{m}$. In the right image, the bounding box of the mask mean $\overline{D}_{m_{\mathrm{box}}}$ is drawn in green, and the bounding box mean $\overline{D}_{b}$ in red.}
            \end{subfigure}
            \vfill
            \begin{subfigure}[h]{0.45\textwidth}
                \centering
                \includegraphics[width=0.45\textwidth,trim={20cm 0 40cm 0},clip]{images/obj_0_heatmap.png}
                \includegraphics[width=0.45\textwidth,trim={40cm 0 20cm 0},clip]{images/obj_0_heatmap.png}
                \caption{The left side shows the mean value of the masks as a heat map before the binary mask was generated. The right shows the standard deviation of the mask also in a heat map representation. }
            \end{subfigure}
            \caption{Instance mask uncertainty example.}
            \label{fig:mask_example}
            \vspace{-10pt}
        \end{figure}
        Besides the bounding box, the used detection model also provides instance masks, which we use to analyze the spatial uncertainty. In Fig.~\ref{fig:mask_example}, we see an example of the mask mean $\overline{D}_{m}$ (top left). Besides the mean, we also calculate the mask's standard deviation $\sigma_{m}$, which is depicted in the bottom right of Fig.~\ref{fig:mask_example}. The heat map of the standard deviation $\sigma_{m}$ reveals that the model well recognizes the left side of the driver's cab since the standard deviation $\sigma_{m}$ is small and represented by a thin line. There is greater uncertainty in the vicinity of the truck radiator as the area is also slightly colored. However, the model shows the largest uncertainty in the transition from truck to trailer, which the intense coloring indicates. 
        
        We need some criteria for the instance mask to use this knowledge. The first criterion reuses the bounding box uncertainty criterion described in Eq.~\ref{eq:box_mean_std}. To calculate the uncertainty for the instance masks related to the individual edges, the bounding box surrounding the mask must be determined with $m^{\mathrm{box}}_{j} = \{c_x,c_y,w,h\} = size(m_{j})$. The mean instance mask $\overline{D}_{m}$ of the object detection is calculated and converted to binary, by
        \begin{equation}\label{eq:mask_mean}
            \overline{D}_{m} =
            \begin{cases}
                1 & \quad \text{if } \frac{\sum^{N}_{j=1} m_{j}}{N} > 0.5\\
                0 & \quad \text{otherwise}
            \end{cases}.
        \end{equation}
        We obtain the mean value of the mask's bounding box with $\overline{D}_{m_{\mathrm{box}}} = \{c_x,c_y,w,h\} = size(\overline{D}_{m})$ and the standard deviation $\sigma_{m_{\mathrm{box}}}$ with Eq.~\ref{eq:mask_std}, which can be used to analyze the edges of the mask bounding box. As with the bounding box in Section~\ref{subsec:box_criteria}, we normalize the standard deviation $\sigma_{m_{\mathrm{box}}}$ based on the mean mask bounding box size. 
        \begin{equation}\label{eq:mask_std}
            \sigma_{m_{\mathrm{box}}}=\sqrt{\frac{\sum^{N}_{j=1} (m^{\mathrm{box}}_{j} - \overline{D}_{m_{\mathrm{box}}})^2}{N-1}}
        \end{equation}
        
        The next criterion we consider appears in \cite{Morrison2019} for result evaluation purposes and is comparable to our bounding box criterion Eq.~\ref{eq:box_iou_mean} already listed above. Besides Eq.~\ref{eq:box_iou_mean}, we also adapt Eq.~\ref{eq:box_iou_std}, where the $\mathrm{IoU}(m_{j}, \overline{D}_{m})$ calculation is slightly changed and results from the mean instance mask $\overline{D}_{m}$ and all other instance masks $m_{j}$. As with the bounding box criterion in Section~\ref{subsec:box_criteria}, the mean $\overline{iou}_m$ and the standard deviation $\sigma_{iou_m}$ provide essential information about the similarity of the masks in the cluster and their spatial uncertainty with respect to the model prediction.
        
        The first mask criterion interprets the instance mask like a bounding box, which reduces the information when considering the uncertainty. Therefore, we propose another criterion to capture the area of the binary instance masks better. To do this, we first calculate the area for each instance mask $A_j=\sum^{H\times W} m_{j}$ and then the mean and standard deviation with Eq.~\ref{eq:mask_area_std}. We then normalize $\sigma_{A_m}$ by the mean $\overline{A}_m$. The standard deviation $\sigma_{A_m}$ provides information about the variance in the size of the mask. A small value is good because all instance masks are similar in size. However, $\sigma_{A_m}$ is not decisive whether they lie spatially together, but this knowledge provides the mask IoU criterion $\overline{iou}_m$ and $\sigma_{iou_m}$.
        \begin{equation}\label{eq:mask_area_std}
            \overline{A}_m = \frac{\sum^{N}_{j=1} A_j}{N},~~\sigma_{A_m}=\sqrt{\frac{\sum^{N}_{j=1} (A_j - \overline{A}_m)^2}{N-1}}
        \end{equation}
 
    \subsection{Bounding Box \& Mask Criteria}\label{subsec:box_mask_criteria}
        \begin{figure}[t]
            \vspace{5pt}
            \centering
            \includegraphics[width=0.475\textwidth,trim={0 0 0 10pt},clip]{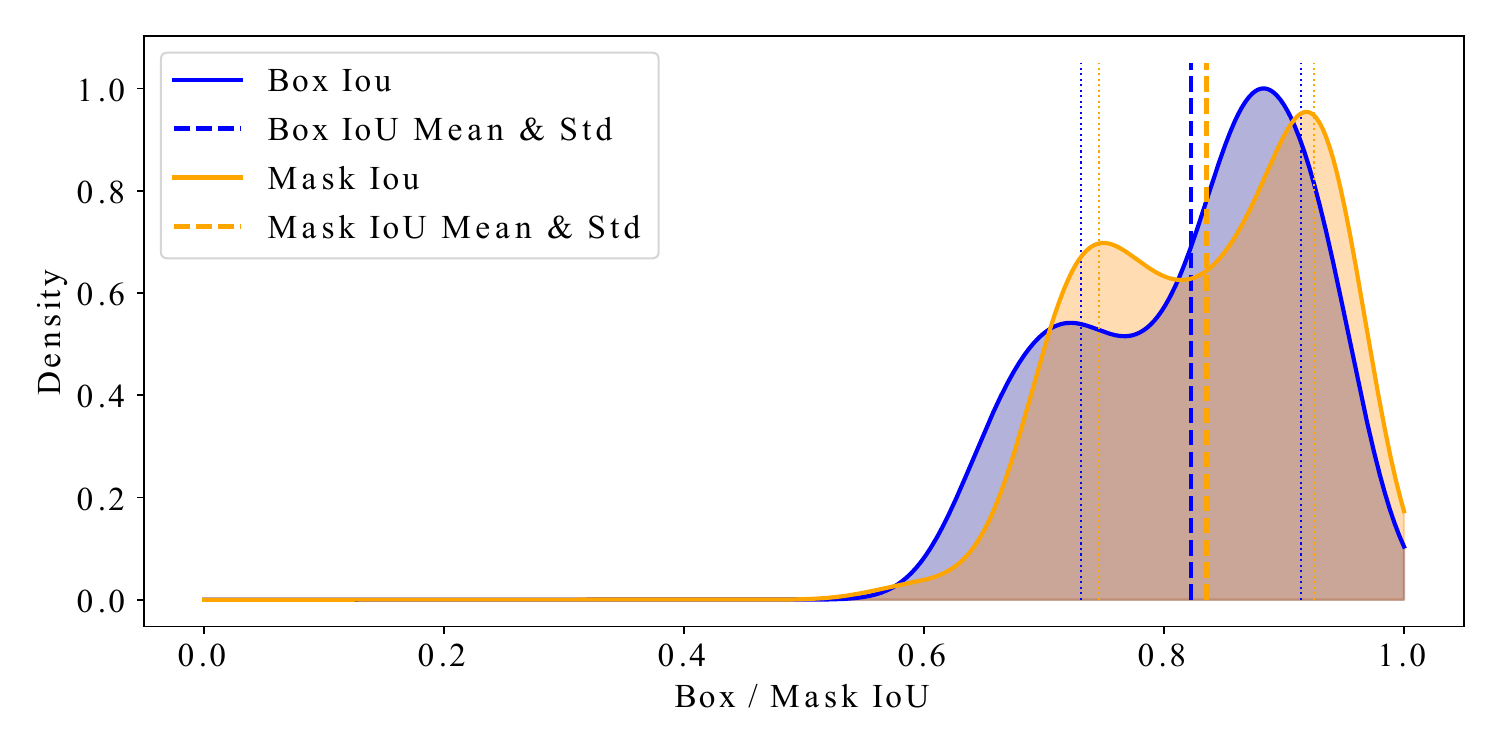}
            \caption{Distribution of the IoU between the mean bounding box $\overline{D}_{b}$ (blue) or mean mask $\overline{D}_{m}$ (orange) and all other bounding box $b_{j}$ or instance mask $m_{j}$.}
            \label{fig:dis_example}
            \vspace{-10pt}
        \end{figure}
        Finally, we present criteria that combine the bounding box and instance mask uncertainty. The assumption for these criteria is that a well-trained ML model provides predictions where the bounding box and instance mask match very well and increase if not. In other words, they match regarding location and size. The first criterion covers the mismatch in terms of IoU $iou_{\mathrm{mis}}$ between the mean bounding box $\overline{D}_{b}$ and the mean bounding box enclosing the instant mask $\overline{D}_{m_{\mathrm{box}}}$ by calculating the IoU between both. An example is depicted in Fig.~\ref{fig:mask_example} (top right). A high IoU value indicates that both matching well.
        \begin{equation}
            iou_{\mathrm{mis}} = \mathrm{IoU}(\overline{D}_{b}, \overline{D}_{m_{\mathrm{box}}})
        \end{equation}
        The first criterion only considers the similarity of the mean values but not the distribution of the predicted bounding boxes or instance masks. However, to compare both distributions, we must first calculate the IoU between the mean and each predicted bounding box $iou_{b_j}$ and instance mask $iou_{m_j}$ within the cluster. If the distribution of the bounding boxes and instance masks is equal, then the two model heads are identical in their predictions, even if the variance of the distribution is large. If the distribution is unequal, the predictions are unequal as the two model heads produce contradictory predictions.
        \begin{equation}
            iou_{b_j} = \mathrm{IoU}(b_{j}, \overline{D}_{b}),~~iou_{m_j} = \mathrm{IoU}(m_{j}, \overline{D}_{m})
        \end{equation}
        To obtain the IoU distribution, we use the kernel density estimation Eq.~\ref{eq:kde} with a Gaussian kernel and compute the probability distribution function with Eq.~\ref{eq:pdf} \cite{Bishop2006}. An example of the IoU distribution for the bounding box and instance mask is shown in Fig.~\ref{fig:dis_example}. The two distributions of the bounding box and mask are not identical. In addition, the distribution reveals the uncertainty arising from the instance mask Fig.~\ref{fig:mask_example} and bounding box Fig.~\ref{fig:bbox_example}. Afterward, we obtain the probability distribution function of the IoU kernel density estimation for the bounding box $p_b=f(iou_{b_j})$ and the instance mask $p_m=f(iou_{m_j})$.
        \begin{equation}\label{eq:kde}
            KDE(x,t) = \frac{1}{n}\overset{i=n}{\underset{i=1}{\sum}} \frac{1}{h\sqrt{2\pi}}\exp^{-0.5(\frac{x_i-t}{h})^2}
        \end{equation}
        \begin{equation}\label{eq:pdf}
            f(t) = \int^1_0 KDE(x,t)\; \mathrm{d}x
        \end{equation}
        With both distributions at hand, we can calculate the Kullback-Leibler-Divergence (KL)~\cite{Bishop2006} to compare the bounding box distribution with the mask distribution $\mathrm{KL}(p_b|p_m)$ and the opposite $\mathrm{KL}(p_m|p_b)$. Besides, we also calculated the Jensen-Shannon Distance (JS)~\cite{Nielsen2020} $\mathrm{JS}(p_b|p_m)$ in Eq.~\ref{eq:js} and the Wasserstein Distance~\cite{Levina2001} $\mathrm{EMD}(p_b|p_m)$ in Eq.~\ref{eq:emd}, also called Earth Mover Distance, to compare both distributions.
        While the JS can be interpreted as merging and averaging the total KL divergence to the average distribution of $\frac{p+q}{2}$, the EMD is more complex.
        \begin{equation}\label{eq:js}
            \mathrm{JS}(p,q)=\sqrt{\frac{\mathrm{KL}(p|m)+\mathrm{KL}(q|m)}{2}},~~m=\frac{p+q}{2}
        \end{equation}
        While $d_{ij}$ describes the distance between $p_i$ and $q_i$, $f_{ij}$ represents the flow. The flow between $p_i$ and $q_i$ has to be minimized to reduce the overall costs with $\sum^m_{i=1}\sum^n_{j=1} f_{ij} = min(\sum^m_{i=1} p_i \sum^n_{j=1} q_j)$. If the optimal flow is found, the EMD~\cite{Levina2001} is defined as
        \begin{equation}\label{eq:emd}
            \mathrm{EMD}(p,q) = \frac{\sum^m_{i=1}\sum^n_{j=1}f_{ij}d_{ij}}{\sum^m_{i=1}\sum^n_{j=1}f_{ij}}.
        \end{equation}

    \subsection{Criteria Evaluation}\label{subsec:eval}
        Next, we examine the value of our proposed uncertainty-based corner case criteria. To this end, we performed forward and backward sequential feature selection\footnote{sklearn.feature\_selection.SequentialFeatureSelector} in combination with a decision tree to determine the top 10 corner case criteria for the classification task. 
        
        Top 10 features selected by forward sequential selection:
            \begin{itemize}
                    \item class score criteria: $\sigma_{c^{k_{\mathrm{max}}}}$, $\overline{D}_{c^{k_{\mathrm{2nd}}}}$, and $\sigma_{c^{k_{\mathrm{2nd}}}}$
                    \item bounding box criteria: $x_1, x_2, w$ of $\sigma_{b}$, $\overline{iou_b}$, and $\sigma_{iou_b}$
                    \item mask criteria: $h$ of $\sigma_{m}$
                    \item bounding box \& mask criteria: $\mathrm{KL}(p_b|p_m)$
            \end{itemize}
            
        Top 10 features selected by backward sequential selection: 
            \begin{itemize}
                \item class score criteria: $\sigma_{c^{k_{\mathrm{max}}}}$, $\overline{D}_{c^{k_{\mathrm{2nd}}}}$, and $\sigma_{c^{k_{\mathrm{2nd}}}}$
                \item bounding box criteria: $c_x$ of $\sigma_{b}$, and $\overline{iou_b}$
                \item mask criteria: $c_x$ of $\sigma_{m}$, and $\sigma_{A_m}$
                \item bounding box \& mask criteria: $iou_{\mathrm{mis}}$, $\mathrm{KL}(p_m|p_b)$, $\mathrm{JS}$
            \end{itemize}

        Furthermore, we evaluated the Pearson and Spearman correlation of bounding box criteria and box IoU, instance mask criteria and mask IoU, and bounding box \& mask criteria concerning box IoU and mask IoU (cf. Supplementary Material, Section~\ref{sec:sup_eval}). The correlations were very low except for some corner case criteria, which have better results for the COCO and the NuImages dataset. These include $\overline{iou_b}$ and $iou_{\mathrm{mis}}$ in correlation with the box IoU, as well as $\overline{iou_m}$, $\sigma_{iou_m}$, $iou_{\mathrm{mis}}$ and $\mathrm{EMD}$ in correlation with the mask IoU. Overall, the mask criteria and bounding box \& mask criteria showed better correlation than the bounding box criteria. However, in Section~\ref{sec:decision}, we used all features for classification because the performance was worse when using any subset of corner case criteria. This can also be assumed due to the poor correlation, which requires all available information to be considered when classifying the objects.

\section{Experimental Setup}\label{sec:exper}
    \begin{table*}[t]
        \vspace{5pt}
        \footnotesize
        \centering
        \def\arraystretch{1.2}
        \begin{tabular}{lll|l|l|l|l|l} \toprule
            \multicolumn{2}{l}{Mask R-CNN with} & no NMS & $N_i=10$ & $N_i=100$ & no NMS & $N_i=10$ & $N_i=100$ \\ \midrule
            \multicolumn{2}{l}{Dataset} & \multicolumn{3}{c|}{COCO}  & \multicolumn{3}{c}{NuImages} \\
            \multicolumn{2}{l}{Size of Test Subset (Images)} & \multicolumn{3}{c|}{4952} & \multicolumn{3}{c}{14884} \\ 
            \multicolumn{2}{l}{Labeled Objects (Test Subset)} & \multicolumn{3}{c|}{36335} & \multicolumn{3}{c}{136074} \\ \midrule
            \multicolumn{2}{l}{Model Predictions} & 34827 & 67569 & 73344 & 106133 & 180557 & 197743 \\ 
            \multicolumn{1}{c}{TP} & 
                    \begin{tabular}[c]{@{}l@{}}Correct Class \& \\ 
                                               $\mathrm{IoU}>0.5$\end{tabular} & 
                    \begin{tabular}[c]{@{}l@{}}16040~~(46.1\%)\end{tabular} &
                    \begin{tabular}[c]{@{}l@{}}23778~~(35.2\%)\end{tabular} &
                    \begin{tabular}[c]{@{}l@{}}24554~~(33.5\%)\end{tabular} &
                    \begin{tabular}[c]{@{}l@{}}59921~~(56.5\%)\end{tabular} &
                    \begin{tabular}[c]{@{}l@{}}100309~(55.6\%)\end{tabular} &
                    \begin{tabular}[c]{@{}l@{}}106441~(53.8\%)\end{tabular} \\ 
            \multicolumn{1}{c}{L-CC} & 
                    \begin{tabular}[c]{@{}l@{}}Correct Class \& \\ 
                                               $0.5>\mathrm{IoU}>0.1$\end{tabular} & 
                    \begin{tabular}[c]{@{}l@{}}3378~~~~(9.7\%)\end{tabular} &
                    \begin{tabular}[c]{@{}l@{}}3213~~~~(4.8\%)\end{tabular} &
                    \begin{tabular}[c]{@{}l@{}}2901~~~~(4.0\%)\end{tabular} & 
                    \begin{tabular}[c]{@{}l@{}}11667~~(11.0\%)\end{tabular} &
                    \begin{tabular}[c]{@{}l@{}}9800~~~~(5.4\%)\end{tabular} &
                    \begin{tabular}[c]{@{}l@{}}7546~~~~(3.8\%)\end{tabular} \\  
            \multicolumn{1}{c}{C-CC} & 
                    \begin{tabular}[c]{@{}l@{}}Wrong Class \& \\ 
                                               $\mathrm{IoU}>0.5$\end{tabular} & 
                    \begin{tabular}[c]{@{}l@{}}1275~~~~(3.7\%)\end{tabular} &
                    \begin{tabular}[c]{@{}l@{}}1714~~~~(2.5\%)\end{tabular} &
                    \begin{tabular}[c]{@{}l@{}}1749~~~~(2.4\%)\end{tabular} & 
                    \begin{tabular}[c]{@{}l@{}}3206~~~~(3.0\%)\end{tabular} &
                    \begin{tabular}[c]{@{}l@{}}4373~~~~(2.4\%)\end{tabular} &
                    \begin{tabular}[c]{@{}l@{}}4429~~~~(2.2\%)\end{tabular} \\  
            \multicolumn{1}{c}{LC-CC} & 
                    \begin{tabular}[c]{@{}l@{}}Wrong Class \& \\ 
                                               $0.5>\mathrm{IoU}>0.1$\end{tabular} & 
                    \begin{tabular}[c]{@{}l@{}}1077~~~~(3.1\%)\end{tabular} &
                    \begin{tabular}[c]{@{}l@{}}1189~~~~(1.8\%)\end{tabular} &
                    \begin{tabular}[c]{@{}l@{}}1205~~~~(1.6\%)\end{tabular} & 
                    \begin{tabular}[c]{@{}l@{}}2931~~~~(2.8\%)\end{tabular} &
                    \begin{tabular}[c]{@{}l@{}}3050~~~~(1.7\%)\end{tabular} &
                    \begin{tabular}[c]{@{}l@{}}2893~~~~(1.5\%)\end{tabular} \\ 
            \multicolumn{1}{l}{FP} & No Matches & 
                    \begin{tabular}[c]{@{}l@{}}13057~~(37.5\%)\end{tabular} &
                    \begin{tabular}[c]{@{}l@{}}37675~~(55.8\%)\end{tabular} &
                    \begin{tabular}[c]{@{}l@{}}42935~~(58.5\%)\end{tabular} & 
                    \begin{tabular}[c]{@{}l@{}}28408~~(26.8\%)\end{tabular} &
                    \begin{tabular}[c]{@{}l@{}}63025~~(34.9\%)\end{tabular} &
                    \begin{tabular}[c]{@{}l@{}}76434~~(38.7\%)\end{tabular} \\ 
            \multicolumn{1}{l}{FN} & Missed Objects & 14565 & 6441 & 5926 & 58349 & 18542 & 14765 \\ \midrule
            \multicolumn{2}{l}{Bounding Box $mAP_{\mathrm{IoU} > 0.5}$} & 0.374 & 0.488 & 0.504 & 0.266 & 0.414 & 0.439 \\ 
            \multicolumn{2}{l}{Mask $mAP_{\mathrm{IoU} > 0.5}$}         & 0.372 & 0.470 & 0.484 & 0.245 & 0.357 & 0.376 \\
            \bottomrule 
        \end{tabular}
        \caption{Test dataset details, number of matching predictions, corner cases, and model performance.}\label{tab:details}
        \vspace{-10pt}
    \end{table*}
    As mentioned in Section~\ref{sec:over}, we used a modified Mask R-CNN model~\cite{Heidecker2023, Morrison2019, Heidecker2021a, Riedlinger2022}. The major difference in comparison to the original Mask R-CNN model from He~et~al.~\cite{He2017} are the added MC-Dropout~\cite{Gal2016} layers to sample from the model, which allows us to model the predictive uncertainty~\cite{Huellermeier2021}. We used a dropout rate of $0.2$ and did $R=10$ and $R=100$ repetitions for each input image to see the difference in the results, as it takes 10 times longer to do 100 compared to 10 repetitions. In the best case of 100 repetitions, we get 400 predictions if the input images contain, for example, four objects. To cluster these predictions, we used Bayesian Gaussian Mixture~\cite{Blei2006} in the post-process, and as features, the bounding box parameters $b = (x_1, y_1, x_2, y_2)$ as described in \cite{Heidecker2023}. We get four object detections $D_i$ containing $N_i=100$ predictions each under ideal conditions. For comparison reasons, we also used the original Mask R-CNN model and disabled Non-maximum Suppression (NMS) as proposed in \cite{Hall2020, Feng2021, Harakeh2019} to get, in the best case, multiple predictions 
    of each object to cluster and determine the corner case criteria.

    For the evaluation, we selected the COCO~\cite{Lin2015} and NuImages~\cite{Caesar2020} datasets to analyze our corner case criteria because both datasets are very different. For example, the NuImages test subset is around three times bigger than COCO, as listed in Table~\ref{tab:details}. In both cases, we are using the official validation subset as a test subset to evaluate the corner case criteria and the overall approach. Furthermore, we randomly picked $10\%$ from the training subset to get a new validation split because both datasets offer only two labeled subsets. With the rest of the training subset, we trained the model. In the case of the COCO dataset, we trained the model with all 81 provided classes. However, in the NuImages dataset, we trained the model with the following 17 classes \textit{(animal, child, pedestrian, other pedestrians (e.g., police), e-scooter, stroller, barrier, traffic cone, debris, bicycle, motorcycle, car, bus, truck, trailer, off-highway, other vehicles (e.g., police))}. For the assignment of detection to the corner case categories defined in Section~\ref{subsec:cc_cat}, we use the Hungarian Matching Algorithm~\cite{Stewart2016} combined with the class to identify the matches between GT and the predicted object detections $D_i$. 
    
    As mentioned before, both datasets are quite different. Also, the mAP of the model performance differs; on the COCO~\cite{Lin2015} dataset, the model is better than on the NuImages~\cite{Caesar2020} dataset, which is surprising because COCO has 81 classes and NuImage 17. Since the model without NMS misses many objects, the number of predicted objects is also quite low compared to the MC-Dropout model. Another expected behavior also occurs, namely that more objects are predicted as repetitions increase. The model ideally manages to predict every object in the image at least once during the repetition, which in turn also leads to objects being predicted where there are none (No Matches in Table~\ref{tab:details}). It is also interesting to note that the predictions classified as corner cases have a similar quantity of objects. Our current assumption for the MC-Dropout models is that the contained objects overlap to a high degree, i.e., objects representing a corner case at a repetition of 10 also represent one at 100 repetitions.

\section{Corner Case Decision Function}\label{sec:decision}
    With the criteria shown in Section~\ref{sec:criteria}, we have a proper baseline to quantify how certain or uncertain the model is with respect to its prediction about each object. It should also be noted that we do not rely on GT data to calculate the corner case criteria, apart from the training of the object detection model in Section~\ref{sec:exper}. The question is, can we classify the corner case categories from the overview in Section~\ref{sec:criteria}? -- Yes, to some extent. 

    A "corner case" decision function is needed to achieve this goal. For this purpose, we used five different classification methods from sklearn~\cite{Pedregosa2011}: Decision Tree (DT), Random Forest (RF), Support Vector Machine (SVM), Gaussian Process Classifier (GPC), and a Multi-layer Perceptron (MLP) classifier with only one hidden layer (100 neurons). Besides that, all default parameter settings were kept.

    \begin{table}[ht]
        \centering
        \begin{tabular}{lc|c|c|c|c|c}
            \toprule
            \begin{tabular}[c]{@{}l@{}}Mask \\ R-CNN \end{tabular} & 
            \rotatebox[origin=c]{90}{no NMS} & \rotatebox[origin=c]{90}{$N_i=10$} & \rotatebox[origin=c]{90}{$N_i=100$} & \rotatebox[origin=c]{90}{no NMS} & \rotatebox[origin=c]{90}{$N_i=10$} & \rotatebox[origin=c]{90}{$N_i=100$} \\ 
            Dataset & \multicolumn{3}{c|}{COCO}  & \multicolumn{3}{c}{NuImages} \\ \midrule
            DT              & 0.56 & \textbf{0.5 } & 0.43 & 0.56 & \textbf{0.5 } & 0.43 \\ \midrule
            RF              & 0.5  & 0.38 & 0.39 & 0.5  & 0.38 & 0.39 \\ \midrule
            SVM             & 0.43 & 0.31 & 0.28 & 0.42 & 0.31 & 0.28 \\ \midrule
            GPC             & \textbf{0.58} & \textbf{0.5} & \textbf{0.45} & \textbf{0.58} & \textbf{0.5 } & \textbf{0.45} \\ \midrule
            MLP             & 0.56 & 0.44 & 0.4  & 0.56 & 0.44 & 0.4  \\
            \bottomrule 
        \end{tabular}
        \caption{Class-weighted F1-score results using Decision Tree (DT), Random Forest (RF), Support Vector Machine (SVM), Gaussian Process Classifier (GPC) and Multi-layer Perception (MLP). Input features are the corner cases criteria (cf. Section~\ref{sec:criteria}), derived from the mentioned Mask R-CNN model and the class labels provided by the corner case categories (cf. Section~\ref{subsec:cc_cat}).}
        \label{tab:f1}
        \vspace{-8pt}
    \end{table}
    
    As input feature, we used all presented corner case criteria, and as output, the five classes: TP, L-CC, C-CC, LC-CC, and FP (cf. Table~\ref{tab:details}). We trained each classifier model with the validation dataset, but we had to use Random Undersampling~\cite{Guillaume2017} because of the class imbalance. Table~\ref{tab:f1} reveals that the Gaussian Process Classifier outperforms all other methods. Interestingly, the classification results for the corner case criteria of the Mask R-CNN model with disabled NMS are much better than for those models where the corner case criteria were calculated on the MC-Dropout model results. Thus, there seems to be a connection that many object predictions (cf. Table~\ref{tab:details}), which mainly concern non-existent objects, have a negative effect on the classification, e.g., COCO with a repetition of 10 and 100.

    \begin{figure}[ht]
        \centering
        \includegraphics[width=0.4\textwidth, trim={0 20pt 0 25pt},clip]{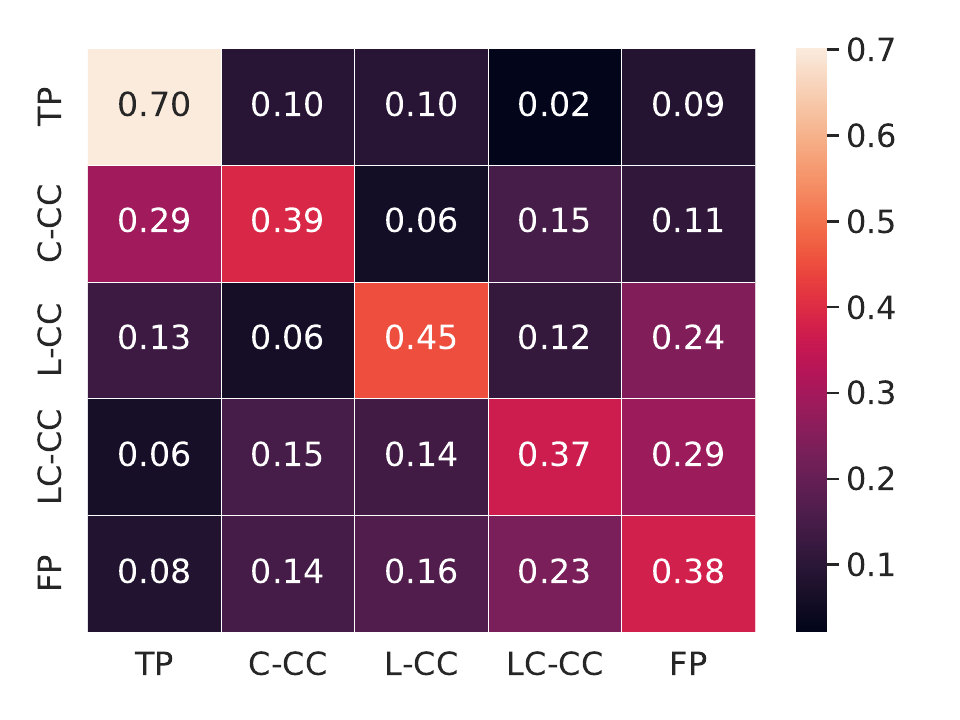}
        \caption{The Confusion matrix of our decision function, using a Gaussian Process Classifier ($N_i=10$) and presented corner cases criteria as input, to distinguish objects in TP, corner case, or FP.}
        \label{fig:confusion_matrix}
        \vspace{-8pt}
    \end{figure}

    The class-weighted F1-score of the Gaussian Process Classifier ($N_i=10$) decision function is $0.5$, and the related confusion matrix in Fig.~\ref{fig:confusion_matrix} shows further details. A breakdown of the corner cases into individual classes is possible to a certain extent. The L-CC and LC-CC corner cases are challenging to separate from FP, which is not surprising since FP contains object detections whose IoU is only minimally smaller than $0.1$ and would otherwise perhaps represent a corner case. Furthermore, FP contains duplicates of objects from other categories. This happens if there are several matches for a GT label, but only the best match is selected as a corner case or TP.

\section{Corner Case Detection for Data Reduction}\label{sec:iter_retrain_model}
    In this section, we apply our approach, as shown in Fig.~\ref{fig:approach}, to a real scenario where each data and annotation costs money. Therefore, we evaluate using our corner case criteria to reduce the amount of training data so that we need fewer annotations but still achieve good results compared to the baseline model trained with the entire training dataset. We perform multiple training cycles to improve the model iteratively by adding model-selected corner case data to the training dataset to achieve this goal. We split the NuImages training dataset into four equal-sized subsets. In the 1st cycle, we train the model on the 1st training data subset and execute the inference and corner case decision on the 2nd training data subset. Every image that contains at least one corner case (L-CC, C-CC, or LC-CC) is added to the training of the 2nd cycle, and we repeat the process. 
    
    \begin{figure}[htb]
        \vspace{-8pt}
        \centering
        \includegraphics[width=0.475\textwidth, trim={20pt 0pt 0 0pt},clip]{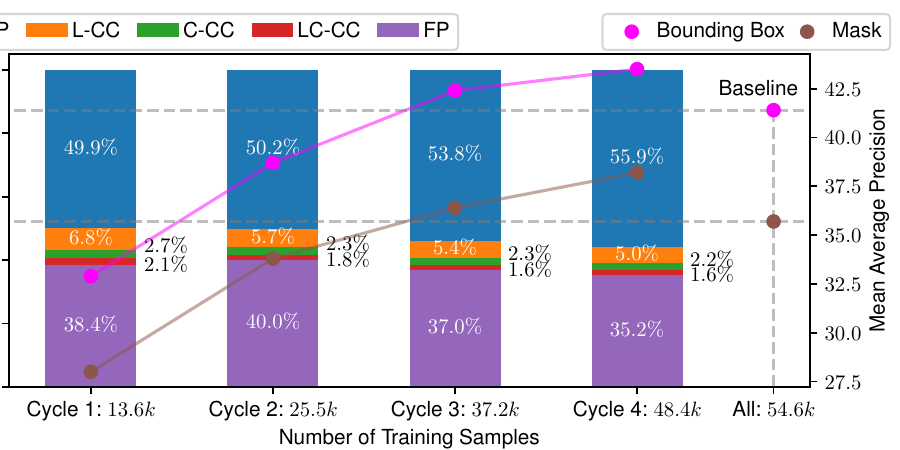}
        \caption{Four training cycles to iteratively improve the model by adding model-selected corner case data to the training dataset. Performance comparison between the cyclically trained model and a baseline model trained with all data.}
        \label{fig:iter_process}
        \vspace{-8pt}
    \end{figure}
    
    As shown in Fig.~\ref{fig:iter_process}, the proportion of TP prediction on the testing dataset increases after each training cycle, and the FN decreases simultaneously. Adding only the model detected corner cases to the training dataset improves the model performance at each cycle iteration. We also see that the number of corner cases (for the model) decreases due to the improvement.
    Since we find a corner case in almost every image, the reduction in training data used after four cycles is moderate, with 6k images (\textasciitilde{$10\%$}) compared to when all data is used. However, our model performs better than the baseline trained on all data. In future works, it would be interesting to investigate the model performance if only images with two or more corner cases were added to the training dataset, and we have maybe a reduction of $30\%$ or more and an even higher cost saving.


\section{Conclusion and Future Work}\label{sec:conclusion}
    The approach presented in this article aims to detect corner cases based on uncertainties and use them for better model training (reduction of annotation costs). Therefore, we introduce corner case criteria derived from the predictive uncertainty of probabilistic object detectors. The corner case criteria determine the uncertainty of the predicted class score, bounding box, and instance mask, which classifies the detected object. It should be noted that the criteria are computed purely from the modeled uncertainty, assuming a sampling predictive distribution with subsequent clustering, where no ground truth (GT) is required. 
    The presented corner case decision function utilizes the output of the corner case criteria as input features and assigns the detected objects to one of the following corner case categories: True Positive (TP), Localization Corner Case (L-CC), Classification Corner Case (C-CC), Localization \& Classification Corner Case (LC-CC), and False Positive (FP). The classification results show we can identify many corner cases based on the introduced criteria. On the other hand, it also shows that the distinction between FP and corner cases is challenging. Finally, we presented our first results on an iterative training cycle that outperforms the baseline model by using the corner cases as a selection strategy for model training to reduce costs regarding annotations.

    
    In the future, we also want to investigate further how the model performance behaves if we only add images to the training dataset containing multiple corner cases and if our results are also statistically consistent.
    With the ability to detect and classify uncertainty-based corner cases, we are also interested in developing relevance metrics to determine the value of the recognized corner cases~\cite{Roesch2022, Breitenstein2023}. To explicitly state how important or valuable a corner case is, e.g., concerning vehicle safety or model training.

\section*{Acknowledgements} This work results from the project KI Data Tooling \linebreak (19A20001O) funded by the German Federal Ministry for Economic Affairs and Climate Action (BMWK).
    
{
    \small
    \bibliographystyle{ieeenat_fullname}
    \bibliography{bibliography}

\begin{thebibliography}{43}
\providecommand{\natexlab}[1]{#1}
\providecommand{\url}[1]{\texttt{#1}}
\expandafter\ifx\csname urlstyle\endcsname\relax
  \providecommand{\doi}[1]{doi: #1}\else
  \providecommand{\doi}{doi: \begingroup \urlstyle{rm}\Url}\fi

\bibitem[Bishop(2006)]{Bishop2006}
Christopher~M. Bishop.
\newblock \emph{{Pattern Recognition and Machine Learning (Information Science
  and Statistics)}}.
\newblock Springer-Verlag, New York, NY, USA, 2006.

\bibitem[Blei and Jordan(2006)]{Blei2006}
David~M. Blei and Michael~I. Jordan.
\newblock {Variational Inference for Dirichlet Process Mixtures}.
\newblock \emph{Bayesian Analysis}, 1\penalty0 (1):\penalty0 121--144, 2006.

\bibitem[Bogdoll et~al.(2021)Bogdoll, Breitenstein, Heidecker, Bieshaar, Sick,
  Fingscheidt, and Zöllner]{Bogdoll2021}
Daniel Bogdoll, Jasmin Breitenstein, Florian Heidecker, Maarten Bieshaar,
  Bernhard Sick, Tim Fingscheidt, and J.~Marius Zöllner.
\newblock {Description of Corner Cases in Automated Driving: Goals and
  Challenges}.
\newblock In \emph{Proc. of ICCV, ERCVAD Workshop}, pages 1023--1028, Virtual,
  2021.

\bibitem[Bolte et~al.(2019)Bolte, B\"{a}r, Lipinski, and
  Fingscheidt]{Bolte2019}
Jan-Aike Bolte, Andreas B\"{a}r, Daniel Lipinski, and Tim Fingscheidt.
\newblock {Towards Corner Case Detection for Autonomous Driving}.
\newblock In \emph{Proc. of IV}, pages 438--445, Paris, France, 2019.

\bibitem[Bolya et~al.(2020)Bolya, Foley, Hays, and Hoffman]{Bolya2020}
Daniel Bolya, Sean Foley, James Hays, and Judy Hoffman.
\newblock {TIDE: A General Toolbox for Identifying Object Detection Errors}.
\newblock In \emph{Proc. of ECCV}, pages 558--573, Virtual, 2020.

\bibitem[Breitenstein et~al.(2023)Breitenstein, Heidecker, Lyssenko, Bogdoll,
  Bieshaar, Zöllner, Sick, and Fingscheidt]{Breitenstein2023}
Jasmin Breitenstein, Florian Heidecker, Maria Lyssenko, Daniel Bogdoll, Maarten
  Bieshaar, J.~Marius Zöllner, Bernhard Sick, and Tim Fingscheidt.
\newblock {What Does Really Count? Estimating Relevance of Corner Cases for
  Semantic Segmentation in Automated Driving}.
\newblock In \emph{Proc. of ICCV, Workshops}, pages 3991--4000, Paris, France,
  2023.

\bibitem[Caesar et~al.(2020)Caesar, Bankiti, Lang, Vora, Liong, Xu, Krishnan,
  Pan, Baldan, and Beijbom]{Caesar2020}
Holger Caesar, Varun Bankiti, Alex~H. Lang, Sourabh Vora, Venice~Erin Liong,
  Qiang Xu, Anush Krishnan, Yu Pan, Giancarlo Baldan, and Oscar Beijbom.
\newblock {nuScenes: A Multimodal Dataset for Autonomous Driving}.
\newblock In \emph{Proc. of CVPR}, pages 11618--11628, Seattle, WA, USA, 2020.

\bibitem[Chan et~al.(2021)Chan, Rottmann, and Gottschalk]{Chan2021}
Robin Chan, Matthias Rottmann, and Hanno Gottschalk.
\newblock {Entropy Maximization and Meta Classification for Out-of-Distribution
  Detection in Semantic Segmentation}.
\newblock In \emph{Proc. of ICCV}, pages 5108--5117, Montreal, QC, Canada,
  2021.

\bibitem[Chandola et~al.(2009)Chandola, Banerjee, and Kumar]{Chandola2009}
Varun Chandola, Arindam Banerjee, and Vipin Kumar.
\newblock {Anomaly Detection: A Survey}.
\newblock \emph{ACM Computing Surveys}, 41\penalty0 (3):\penalty0 1--58, 2009.

\bibitem[Feng et~al.(2021)Feng, Harakeh, Waslander, and Dietmayer]{Feng2021}
Di Feng, Ali Harakeh, Steven Waslander, and Klaus Dietmayer.
\newblock {A Review and Comparative Study on Probabilistic Object Detection in
  Autonomous Driving}.
\newblock \emph{IEEE Trans. on ITS}, pages 1--20, 2021.

\bibitem[Gal and Ghahramani(2016)]{Gal2016}
Yarin Gal and Zoubin Ghahramani.
\newblock {Dropout as a Bayesian Approximation: Representing Model Uncertainty
  in Deep Learning}.
\newblock In \emph{Proc. of ICML}, pages 1050--1059, New York, NY, USA, 2016.

\bibitem[Gawlikowski et~al.(2022)Gawlikowski, Tassi, Ali, Lee, Humt, Feng,
  Kruspe, Triebel, Jung, Roscher, Shahzad, Yang, Bamler, and
  Zhu]{Gawlikowski2022}
Jakob Gawlikowski, Cedrique Rovile~Njieutcheu Tassi, Mohsin Ali, Jongseok Lee,
  Matthias Humt, Jianxiang Feng, Anna Kruspe, Rudolph Triebel, Peter Jung,
  Ribana Roscher, Muhammad Shahzad, Wen Yang, Richard Bamler, and Xiao~Xiang
  Zhu.
\newblock {A Survey of Uncertainty in Deep Neural Networks}.
\newblock \emph{arXiv preprint arXiv:2107.03342}, 2022.

\bibitem[Goodfellow et~al.(2015)Goodfellow, Shlens, and
  Szegedy]{Goodfellow2015}
I. Goodfellow, J. Shlens, and C. Szegedy.
\newblock {Explaining and Harnessing Adversarial Examples}.
\newblock In \emph{Proc. of ICLR}, pages 1--10, San Diego, CA, USA, 2015.

\bibitem[Gruhl et~al.(2021)Gruhl, Sick, and Tomforde]{Gruhl2021}
Christian Gruhl, Bernhard Sick, and Sven Tomforde.
\newblock {Novelty Detection in Continuously Changing Environments}.
\newblock \emph{Future Generation Computer Systems}, 114:\penalty0 138--154,
  2021.

\bibitem[Hall et~al.(2020)Hall, Dayoub, Skinner, Zhang, Miller, Corke,
  Carneiro, Angelova, and S{\"u}nderhauf]{Hall2020}
David Hall, Feras Dayoub, John Skinner, Haoyang Zhang, Dimity Miller, Peter
  Corke, Gustavo Carneiro, Anelia Angelova, and Niko S{\"u}nderhauf.
\newblock {Probabilistic Object Detection: Definition and Evaluation}.
\newblock In \emph{Proc. of WACV}, pages 1031--1040, Snowmass Village, CO, USA,
  2020.

\bibitem[Harakeh et~al.(2019)Harakeh, Smart, and Waslander]{Harakeh2019}
Ali Harakeh, Michael Smart, and Steven~L. Waslander.
\newblock {BayesOD: A Bayesian Approach for Uncertainty Estimation in Deep
  Object Detectors}.
\newblock \emph{arXiv preprint arXiv:1903.03838}, 2019.

\bibitem[Hawkins(1980)]{Hawkins1980}
D.~M. Hawkins.
\newblock \emph{{Identification of Outliers}}.
\newblock Springer, Dordrecht, 1980.

\bibitem[He et~al.(2017)He, Gkioxari, Doll\'ar, and Girshick]{He2017}
Kaiming He, Georgia Gkioxari, Piotr Doll\'ar, and Ross Girshick.
\newblock {Mask R-CNN}.
\newblock In \emph{Proc. of ICCV}, pages 2980--2988, Venice, Italy, 2017.

\bibitem[Heidecker et~al.(2021{\natexlab{a}})Heidecker, Breitenstein, Rösch,
  Löhdefink, Bieshaar, Stiller, Fingscheidt, and Sick]{Heidecker2021b}
Florian Heidecker, Jasmin Breitenstein, Kevin Rösch, Jonas Löhdefink, Maarten
  Bieshaar, Christoph Stiller, Tim Fingscheidt, and Bernhard Sick.
\newblock {An Application-Driven Conceptualization of Corner Cases for
  Perception in Highly Automated Driving}.
\newblock In \emph{Proc. of the IV}, pages 644--651, Nagoya, Japan,
  2021{\natexlab{a}}.

\bibitem[Heidecker et~al.(2021{\natexlab{b}})Heidecker, Hannan, Bieshaar, and
  Sick]{Heidecker2021a}
Florian Heidecker, Abdul Hannan, Maarten Bieshaar, and Bernhard Sick.
\newblock {Towards Corner Case Detection by Modeling the Uncertainty of
  Instance Segmentation Networks}.
\newblock In \emph{Proc. of ICPR, IADS Workshop}, pages 361--374, Milan, Italy,
  2021{\natexlab{b}}.

\bibitem[Heidecker et~al.(2023)Heidecker, El-Khateeb, and Sick]{Heidecker2023}
Florian Heidecker, Ahmad El-Khateeb, and Bernhard Sick.
\newblock {Sampling-based Uncertainty Estimation for an Instance Segmentation
  Network}.
\newblock \emph{arXiv preprint arXiv:2305.14977}, 2023.

\bibitem[Houben et~al.(2021)Houben, Abrecht, Akila, Bär, Brockherde, Feifel,
  Fingscheidt, Gannamaneni, Ghobadi, Hammam, Haselhoff, Hauser, Heinzemann,
  Hoffmann, Kapoor, Kappel, Klingner, Kronenberger, Küppers, Löhdefink,
  Mlynarski, Mock, Mualla, Pavlitskaya, Poretschkin, Pohl, Ravi-Kumar,
  Rosenzweig, Rottmann, Rüping, Sämann, Schneider, Schulz, Schwalbe, Sicking,
  Srivastava, Varghese, Weber, Wirkert, Wirtz, and Woehrle]{Houben2021}
Sebastian Houben, Stephanie Abrecht, Maram Akila, Andreas Bär, Felix
  Brockherde, Patrick Feifel, Tim Fingscheidt, Sujan~Sai Gannamaneni,
  Seyed~Eghbal Ghobadi, Ahmed Hammam, Anselm Haselhoff, Felix Hauser, Christian
  Heinzemann, Marco Hoffmann, Nikhil Kapoor, Falk Kappel, Marvin Klingner, Jan
  Kronenberger, Fabian Küppers, Jonas Löhdefink, Michael Mlynarski, Michael
  Mock, Firas Mualla, Svetlana Pavlitskaya, Maximilian Poretschkin, Alexander
  Pohl, Varun Ravi-Kumar, Julia Rosenzweig, Matthias Rottmann, Stefan Rüping,
  Timo Sämann, Jan~David Schneider, Elena Schulz, Gesina Schwalbe, Joachim
  Sicking, Toshika Srivastava, Serin Varghese, Michael Weber, Sebastian
  Wirkert, Tim Wirtz, and Matthias Woehrle.
\newblock {Inspect, Understand, Overcome: A Survey of Practical Methods for AI
  Safety}.
\newblock \emph{arXiv preprint arXiv:2104.14235}, 2021.

\bibitem[H{\"u}llermeier and Waegeman(2021)]{Huellermeier2021}
Eyke H{\"u}llermeier and Willem Waegeman.
\newblock {Aleatoric and Epistemic Uncertainty in Machine Learning: An
  Introduction to Concepts and Methods}.
\newblock \emph{Machine Learning}, 110\penalty0 (3):\penalty0 457--506, 2021.

\bibitem[Jia et~al.(2022)Jia, Yang, Jia, Zhao, Zhou, and Song]{Jia2022}
Wenhe Jia, Lu Yang, Zilong Jia, Wenyi Zhao, Yilin Zhou, and Qing Song.
\newblock {TIVE: A Toolbox for Identifying Video Instance Segmentation Errors}.
\newblock \emph{arXiv preprint arXiv:2210.08856}, 2022.

\bibitem[Kottke et~al.(2021)Kottke, Herde, Sandrock, Huseljic, Krempl, and
  Sick]{Kottke2021}
Daniel Kottke, Marek Herde, Christoph Sandrock, Denis Huseljic, Georg Krempl,
  and Bernhard Sick.
\newblock {Toward Optimal Probabilistic Active Learning Using a Bayesian
  Approach}.
\newblock \emph{Machine Learning}, 110\penalty0 (6):\penalty0 1199--–1231,
  2021.

\bibitem[Lakshminarayanan et~al.(2017)Lakshminarayanan, Pritzel, and
  Blundell]{Lakshminarayanan2017}
Balaji Lakshminarayanan, Alexander Pritzel, and Charles Blundell.
\newblock {Simple and Scalable Predictive Uncertainty Estimation Using Deep
  Ensembles}.
\newblock In \emph{Proc. of NIPS}, pages 6402--6413, Long Beach, CA, USA, 2017.

\bibitem[Lema{{\^i}}tre et~al.(2017)Lema{{\^i}}tre, Nogueira, and
  Aridas]{Guillaume2017}
Guillaume Lema{{\^i}}tre, Fernando Nogueira, and Christos~K. Aridas.
\newblock {Imbalanced-learn: A Python Toolbox to Tackle the Curse of Imbalanced
  Datasets in Machine Learning}.
\newblock \emph{Journal of Machine Learning Research}, 18\penalty0
  (17):\penalty0 1--5, 2017.

\bibitem[Levina and Bickel(2001)]{Levina2001}
E. Levina and P. Bickel.
\newblock {The Earth Mover's distance is the Mallows distance: some insights
  from statistics}.
\newblock In \emph{Proc. of ICCV}, pages 251--256, Vancouver, BC, Canada, 2001.

\bibitem[Lin et~al.(2015)Lin, Maire, Belongie, Bourdev, Girshick, Hays, Perona,
  Ramanan, Zitnick, and Dollár]{Lin2015}
Tsung-Yi Lin, Michael Maire, Serge Belongie, Lubomir Bourdev, Ross Girshick,
  James Hays, Pietro Perona, Deva Ramanan, C.~Lawrence Zitnick, and Piotr
  Dollár.
\newblock {Microsoft COCO: Common Objects in Context}.
\newblock \emph{arXiv preprint arXiv:1405.0312}, 2015.

\bibitem[Lis et~al.(2019)Lis, Nakka, Fua, and Salzmann]{Lis2019}
Krzysztof Lis, Krishna Nakka, Pascal Fua, and Mathieu Salzmann.
\newblock {Detecting the Unexpected via Image Resynthesis}.
\newblock In \emph{Proc. of ICCV}, pages 2152--2161, Seoul, Korea (South),
  2019.

\bibitem[Morrison et~al.(2019)Morrison, Milan, and Antonakos]{Morrison2019}
Doug Morrison, Anton Milan, and Epameinondas Antonakos.
\newblock {Uncertainty-aware Instance Segmentation using Dropout Sampling},
  2019.
\newblock Accessed 15-Feb.-2019.

\bibitem[Nielsen(2020)]{Nielsen2020}
Frank Nielsen.
\newblock {On a Generalization of the Jensen–Shannon Divergence and the
  Jensen–Shannon Centroid}.
\newblock \emph{Entropy}, 22\penalty0 (2):\penalty0 1--24, 2020.

\bibitem[Nitsch et~al.(2021)Nitsch, Itkina, Senanayake, Nieto, Schmidt,
  Siegwart, Kochenderfer, and Cadena]{Nitsch2021}
Julia Nitsch, Masha Itkina, Ransalu Senanayake, Juan Nieto, Max Schmidt, Roland
  Siegwart, Mykel~J. Kochenderfer, and Cesar Cadena.
\newblock {Out-of-Distribution Detection for Automotive Perception}.
\newblock In \emph{Proc. of ITSC}, pages 2938--2943, Indianapolis, IN, USA,
  2021.

\bibitem[Ouyang et~al.(2021)Ouyang, Marco, Isobe, Asoh, Oiwa, and
  Seo]{Ouyang2021a}
Tinghui Ouyang, Vicent~Sanz Marco, Yoshinao Isobe, Hideki Asoh, Yutaka Oiwa,
  and Yoshiki Seo.
\newblock {Corner Case Data Description and Detection}.
\newblock In \emph{Proc. of ICSE Workshop}, pages 19--26, Madrid, Spain, 2021.

\bibitem[Pedregosa et~al.(2011)Pedregosa, Varoquaux, Gramfort, Michel, Thirion,
  Grisel, Blondel, Prettenhofer, Weiss, Dubourg, Vanderplas, Passos,
  Cournapeau, Brucher, Perrot, and Duchesnay]{Pedregosa2011}
Fabian Pedregosa, Ga{{\"e}}l Varoquaux, Alexandre Gramfort, Vincent Michel,
  Bertrand Thirion, Olivier Grisel, Mathieu Blondel, Peter Prettenhofer, Ron
  Weiss, Vincent Dubourg, Jake Vanderplas, Alexandre Passos, David Cournapeau,
  Matthieu Brucher, Matthieu Perrot, and {{\'E}}douard Duchesnay.
\newblock {Scikit-learn: Machine Learning in {P}ython}.
\newblock \emph{JMLR}, 12\penalty0 (85):\penalty0 2825--2830, 2011.

\bibitem[Pfeil et~al.(2022)Pfeil, Wieland, Michalke, and Theissler]{Pfeil2022}
Jerg Pfeil, Jochen Wieland, Thomas Michalke, and Andreas Theissler.
\newblock {On Why the System Makes the Corner Case: AI-based HolisticAnomaly
  Detection for Autonomous Driving}.
\newblock In \emph{Proc. of IV}, pages 337--344, Achen, Germany, 2022.

\bibitem[Pimentel et~al.(2014)Pimentel, Clifton, Clifton, and
  Tarassenko]{Pimentel2014}
Marco A.~F. Pimentel, David~A. Clifton, Lei Clifton, and Lionel Tarassenko.
\newblock {A Review of Novelty Detection}.
\newblock \emph{Signal Processing}, 99:\penalty0 215--249, 2014.

\bibitem[Riedlinger et~al.(2022)Riedlinger, Schubert, Kahl, and
  Rottmann]{Riedlinger2022}
Tobias Riedlinger, Marius Schubert, Karsten Kahl, and Matthias Rottmann.
\newblock \emph{{Deep Neural Networks and Data for AutomatedDriving:
  Robustness, Uncertainty Quantification, and In-sights Towards Safety}},
  chapter Uncertainty Quantification for ObjectDetection: Output- and
  Gradient-BasedApproaches, pages 251--275.
\newblock Springer International Publishing, Cham, Switzerland, 2022.

\bibitem[Rösch et~al.(2022)Rösch, Heidecker, Truetsch, Kowol, Schicktanz,
  Bieshaar, Sick, and Stiller]{Roesch2022}
Kevin Rösch, Florian Heidecker, Julian Truetsch, Kamil Kowol, Clemens
  Schicktanz, Maarten Bieshaar, Bernhard Sick, and Christoph Stiller.
\newblock {Space, Time, and Interaction: A Taxonomy of Corner Cases in
  Trajectory Datasets for Automated Driving}.
\newblock In \emph{Proc. of SSCI, IEEE CIVTS}, pages 1--8, Singapore, 2022.

\bibitem[Sedlmeier et~al.(2019)Sedlmeier, Gabor, Phan, Belzner, and
  Linnhoff-Popien]{Sedlmeier2019}
Andreas Sedlmeier, Thomas Gabor, Thomy Phan, Lenz Belzner, and Claudia
  Linnhoff-Popien.
\newblock {Uncertainty-Based Out-of-Distribution Detection in Deep
  Reinforcement Learning}.
\newblock \emph{arXiv preprint arXiv:1901.02219}, 2019.

\bibitem[Stewart et~al.(2016)Stewart, Andriluka, and Ng]{Stewart2016}
Russell Stewart, Mykhaylo Andriluka, and Andrew~Y. Ng.
\newblock {End-to-End People Detection in Crowded Scenes}.
\newblock In \emph{Proc. of CVPR}, pages 2325--2333, Las Vegas, NV, USA, 2016.

\bibitem[Stocco et~al.(2020)Stocco, Weiss, Calzana, and Tonella]{Stocco2020}
Andrea Stocco, Michael Weiss, Marco Calzana, and Paolo Tonella.
\newblock {Misbehaviour Prediction for Autonomous Driving Systems}.
\newblock In \emph{Proc. of ICSE}, pages 359--371, Seoul, Korea (South), 2020.

\bibitem[Xia et~al.(2020)Xia, Zhang, Liu, Shen, and Yuille]{Xia2020}
Yingda Xia, Yi Zhang, Fengze Liu, Wei Shen, and Alan~L. Yuille.
\newblock {Synthesize then Compare: Detecting Failures and Anomalies for
  Semantic Segmentation}.
\newblock \emph{arXiv preprint arXiv:2003.08440}, 2020.

\end{thebibliography}
}

\clearpage
\setcounter{page}{1}
\maketitlesupplementary

\section{Overview}
    This supplementary material gives further information about our main work, which has not been mentioned there mainly due to space limitations. 

\section{Evaluation}\label{sec:sup_eval}
    We evaluate the corner case criteria listed in Section~\ref{sec:criteria} by testing them with our experimental setup from Section~\ref{sec:exper}. 
    
    \subsection{Class Score Uncertainty}
        \begin{figure}[t]
            \vspace{5pt}
            \centering
            \begin{subfigure}[h]{0.475\textwidth}
                \centering
                \includegraphics[width=\textwidth]{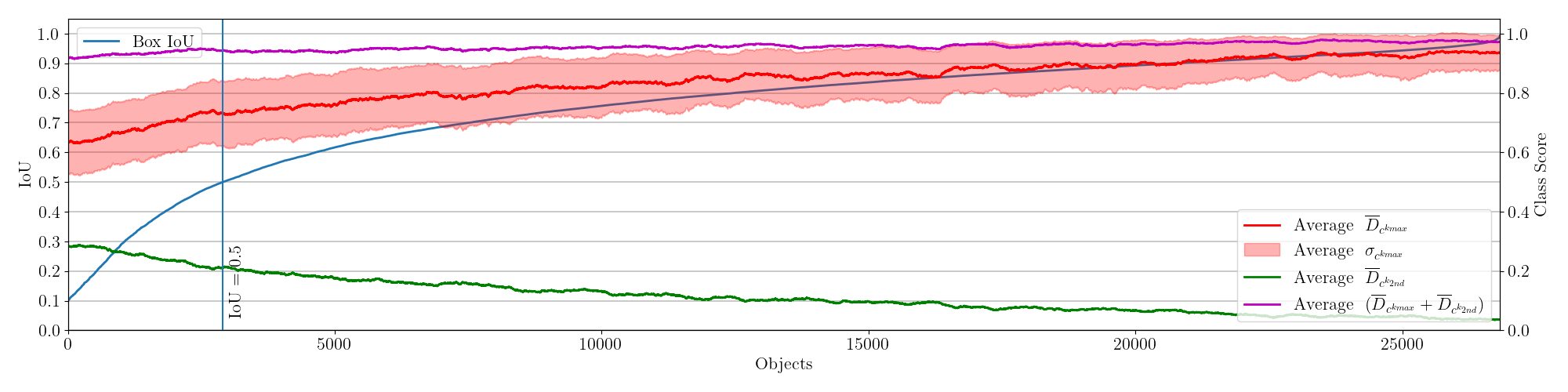}
            \end{subfigure}
            \hfill
            \begin{subfigure}[h]{0.475\textwidth}
                \centering
                \includegraphics[width=\textwidth]{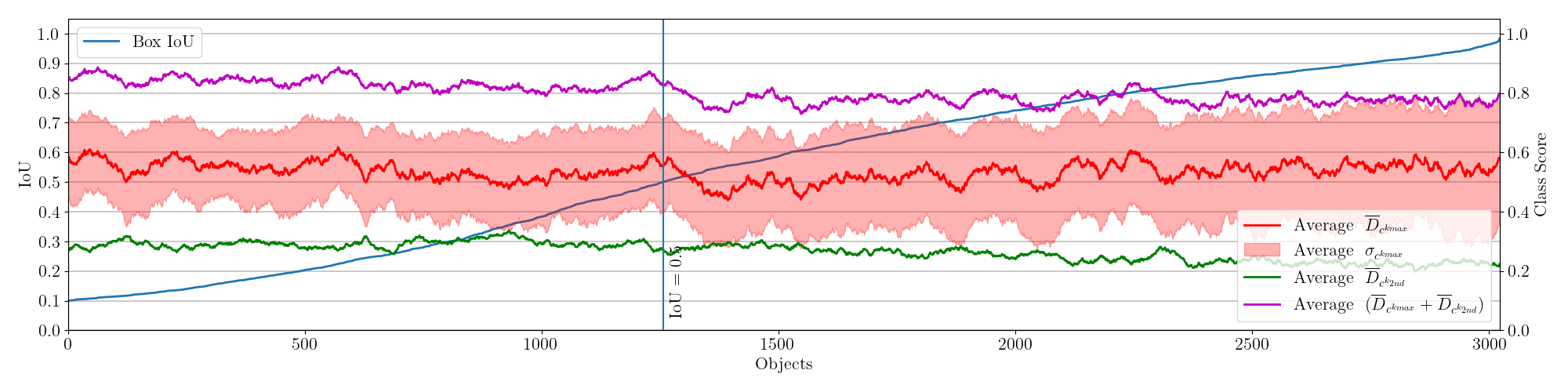}
            \end{subfigure}
            \hfill
            \begin{subfigure}[h]{0.475\textwidth}
                \centering
                \includegraphics[width=\textwidth]{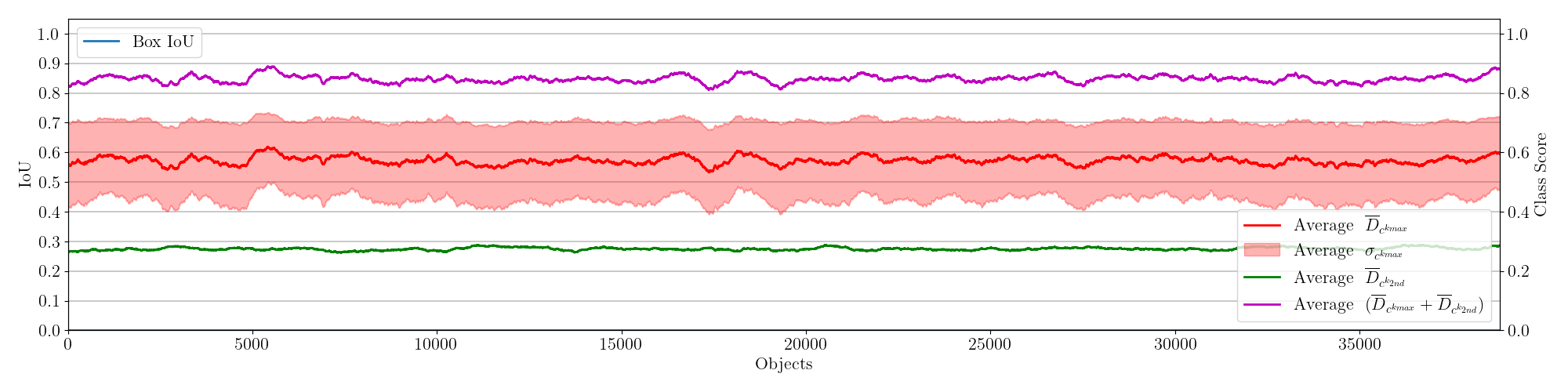}
                \caption{COCO $N_i=100$ results.}
            \end{subfigure}
            \hfill
            \begin{subfigure}[h]{0.475\textwidth}
                \centering
                \includegraphics[width=\textwidth]{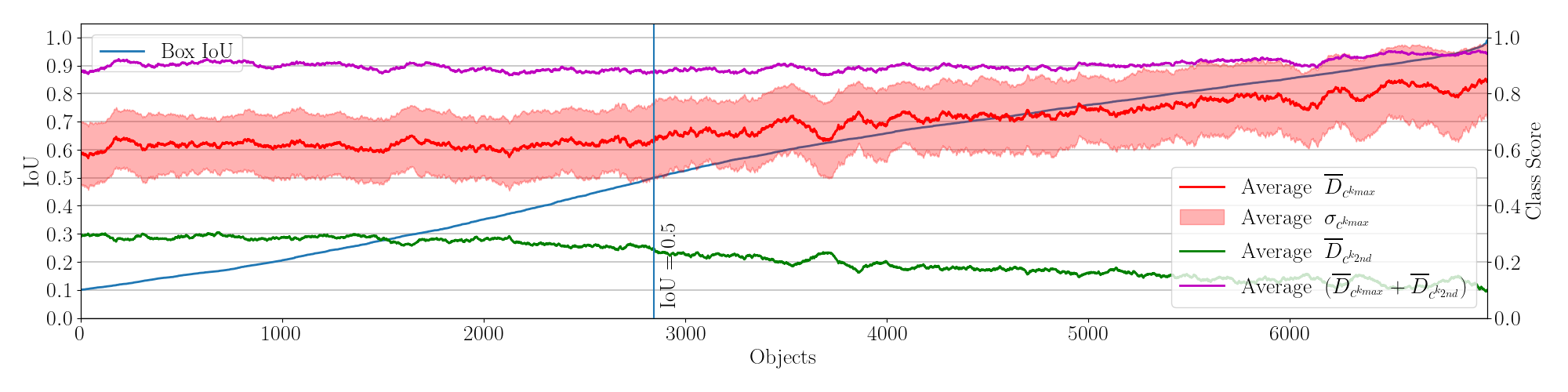}
            \end{subfigure}
            \hfill
            \begin{subfigure}[h]{0.475\textwidth}
                \centering
                \includegraphics[width=\textwidth]{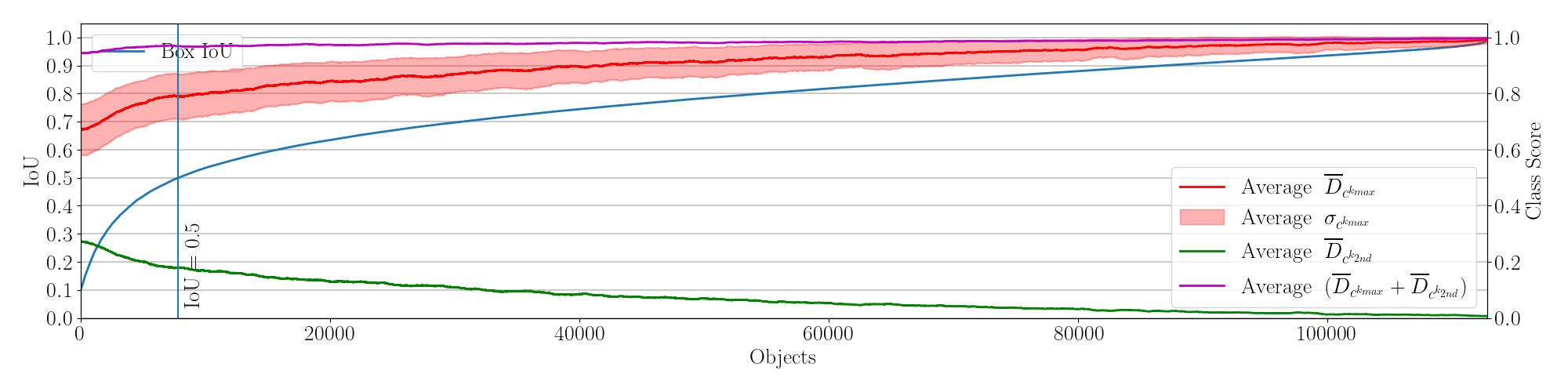}
            \end{subfigure}
            \hfill
            \begin{subfigure}[h]{0.475\textwidth}
                \centering
                \includegraphics[width=\textwidth]{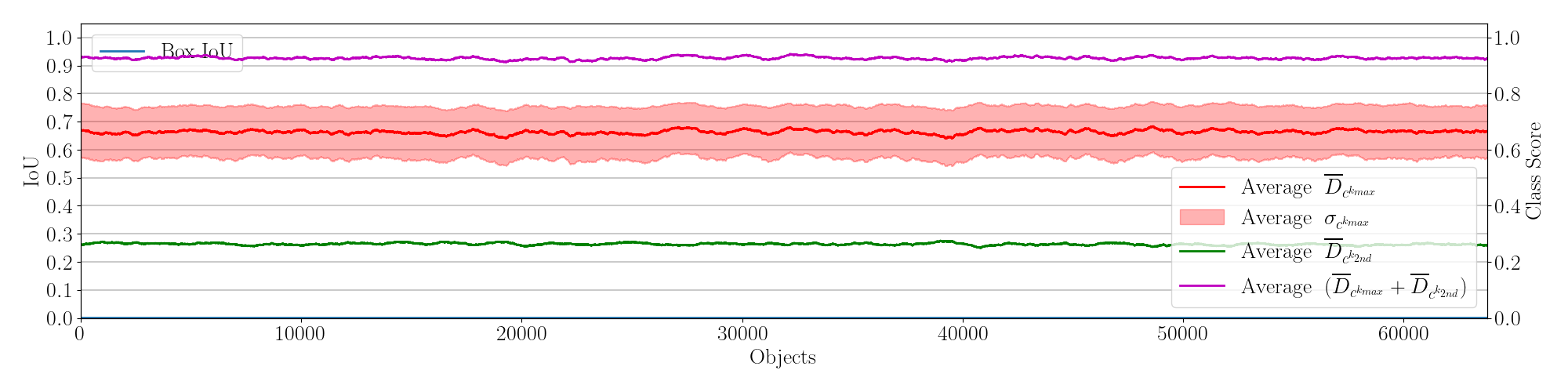}
                \caption{NuImages $N_i=100$ results.}
            \end{subfigure}
            \caption{NuImages results for class score uncertainty. Object evaluation of the max class score mean $\overline{D}_{c^{k_{\mathrm{max}}}}$ (red), 2nd-max class score mean $\overline{D}_{c^{k_{\mathrm{2nd}}}}$ (green), standard deviation $\sigma_{c^{k_{\mathrm{max}}}}$ (red area) and mean sum (magenta).}
            \label{fig:sup_class_criteria}
            \vspace{-10pt}
        \end{figure}
        We depict the results for the NuImages datasets in Fig.~\ref{fig:sup_class_criteria} to evaluate the class score criteria defined in Section~\ref{subsec:class_criteria}. The left y-axis of each plot shows the IoU between mean bounding box $\overline{D}_{b}$ and GT, while the right y-axis represents the class score. The plot on the top contains the cases if the class is correct: \emph{L-CC} and \emph{TP}. The middle presents the cases \emph{LC-CC} and \emph{C-CC}, where the class is wrong, and the last plot finally shows the \emph{FP}. The x-axis represents the predicted objects. Detections are sorted by IoU and smoothed to reveal the trend in class scoring criteria.
        
        We note that the average $\sigma_{c^{k_{\mathrm{max}}}}$ is increasing while the averaged $\overline{D}_{c^{k_{\mathrm{max}}}}$ and IoU is decreasing, see Fig.~\ref{fig:sup_class_criteria} (top). The middle plot behaves unusually because the $\overline{D}_{c^{k_{\mathrm{max}}}}$ is slowly decreasing and then stagnates while the $\sigma_{c^{k_{\mathrm{max}}}}$ is decreasing along with the IoU, which is the opposite of the behavior in the top plot of Fig.~\ref{fig:sup_class_criteria}. In our opinion, this is related to the increase of $\overline{D}_{c^{k_{\mathrm{2nd}}}}$, which is shown in green in Fig.~\ref{fig:sup_class_criteria}. $\overline{D}_{c^{k_{\mathrm{max}}}}$ in the bottom plot forms almost a perfectly straight line with the value corresponding to approx. $0.67$ for NuImages. The value of $\sigma_{c^{k_{\mathrm{max}}}}$ is almost constant. If we take this value and look at the top plot, we see that $\overline{D}_{c^{k_{\mathrm{max}}}}$ is above this value if the class is correct and the standard deviation is the same or smaller. In the middle of Fig.~\ref{fig:sup_class_criteria}, where the class is wrong, $\overline{D}_{c^{k_{\mathrm{max}}}}$ is for most parts below this value, and the standard deviation is the same or larger. This circumstance indicates a threshold to separate correctly and incorrectly classified objects.
        
    \subsection{Spatial Uncertainty}
        \begin{table*}[t]
            \vspace{5pt}
            \footnotesize
            \def\arraystretch{1.2}
            \centering
            \begin{tabular}{|ll|c|c||c|c||c|c||c|c||c|c||c|c|} \hline
                \multicolumn{14}{|l|}{Bounding Box Criteria} \\
                \multicolumn{2}{|l|}{\multirow{2}{*}{Criterion}} & \multicolumn{2}{c||}{COCO} & \multicolumn{2}{c||}{COCO} & \multicolumn{2}{c||}{COCO} & \multicolumn{2}{c||}{NuImages} & \multicolumn{2}{c||}{NuImages} & \multicolumn{2}{c|}{NuImages} \\
                \multicolumn{2}{|l|}{} & \multicolumn{2}{c||}{no NMS} & \multicolumn{2}{c||}{$N_i=10$} & \multicolumn{2}{c||}{$N_i=100$} & \multicolumn{2}{c||}{no NMS} & \multicolumn{2}{c||}{$N_i=10$} & \multicolumn{2}{c|}{$N_i=100$} \\
                \multicolumn{2}{|l|}{} & Pear. & Spear. & Pear. & Spear. & Pear. & Spear. & Pear. & Spear. & Pear. & Spear. & Pear. & Spear. \\ \hline
                $\sigma_{b}$ & $x_1$                     & -0.17 & -0.05 & -0.09 & -0.05 & -0.10 & -0.17 & -0.27 & -0.35 & -0.15 & -0.26 & -0.14 & -0.38 \\ \cline{2-14}
                & $y_1$                                  & -0.16 & -0.09 & -0.10 & -0.11 & -0.13 & -0.25 & -0.21 & -0.29 & -0.12 & -0.14 & -0.16 & -0.29 \\ \cline{2-14}
                & $x_2$                                  & -0.17 & -0.05 & -0.09 & -0.06 & -0.10 & -0.19 & -0.27 & -0.35 & -0.15 & -0.26 & -0.14 & -0.39 \\ \cline{2-14}
                & $y_2$                                  & -0.16 & -0.08 & -0.10 & -0.10 & -0.13 & -0.23 & -0.20 & -0.30 & -0.13 & -0.18 & -0.17 & -0.32 \\ \cline{2-14}
                & $c_x$                                  & -0.17 & -0.08 & -0.09 & -0.08 & -0.10 & -0.21 & -0.27 & -0.37 & -0.15 & -0.29 & -0.14 & -0.41 \\ \cline{2-14}
                & $c_y$                                  & -0.16 & -0.12 & -0.10 & -0.13 & -0.13 & -0.28 & -0.21 & -0.34 & -0.14 & -0.22 & -0.16 & -0.37 \\ \cline{2-14}
                & $w$                                    & -0.13 & -0.06 & -0.08 & -0.07 & -0.16 & -0.21 & -0.21 & -0.31 & -0.16 & -0.23 & -0.25 & -0.40 \\ \cline{2-14}
                & $h$                                    & -0.15 & -0.11 & -0.10 & -0.11 & -0.18 & -0.26 & -0.21 & -0.30 & -0.13 & -0.15 & -0.22 & -0.31 \\ \hline
                \multicolumn{2}{|l|}{$\overline{iou}_b$} & \textbf{ 0.23} &  0.17 & \textbf{ 0.16} &  0.18 & \textbf{ 0.24} & \textbf{ 0.33} & \textbf{ 0.35} & \textbf{ 0.43} &  0.23 &  0.35 & \textbf{ 0.32} & \textbf{ 0.50} \\ \hline
                \multicolumn{2}{|l|}{$\sigma_{iou_b}$}   &  0.02 &  0.06 & -0.02 & -0.03 & -0.14 & -0.19 & -0.01 & -0.10 & -0.09 & -0.17 & -0.21 & -0.33 \\ \hline \hline
    
                \multicolumn{14}{|l|}{Mask Criteria} \\
                $\sigma_{m_{\mathrm{box}}}$ & $c_x$      & \textbf{-0.20} & -0.11 & -0.08 & -0.11 & -0.09 & -0.24 & -0.32 & -0.37 & -0.14 & -0.30 & -0.14 & -0.42 \\ \cline{2-14}
                & $c_y$                                  & -0.16 & -0.13 & -0.09 & -0.15 & -0.13 & \textbf{-0.29} & -0.21 & -0.36 & -0.16 & -0.28 & -0.17 & -0.43 \\ \cline{2-14}
                & $w$                                    & -0.19 & -0.11 & -0.12 & -0.11 & -0.20 & -0.25 & \textbf{-0.34} & -0.33 & \textbf{-0.24} & -0.30 & \textbf{-0.32} & -0.45 \\ \cline{2-14}
                & $h$                                    & \textbf{-0.20} & -0.13 & -0.13 & -0.14 & \textbf{-0.21} & \textbf{-0.29} & -0.32 & -0.34 & -0.19 & -0.25 & -0.25 & -0.41 \\ \hline
                \multicolumn{2}{|l|}{$\overline{iou}_m$} & \textbf{-0.22} & -0.17 & \textbf{-0.15} & \textbf{-0.19} & \textbf{ 0.29} & \textbf{ 0.35} & \textbf{-0.34} & \textbf{-0.38} & \textbf{-0.24} & \textbf{-0.37} & \textbf{ 0.40} & \textbf{ 0.56} \\ \hline
                \multicolumn{2}{|l|}{$\sigma_{iou_m}$}   & \textbf{ 0.31} & \textbf{ 0.22} & \textbf{ 0.21} & \textbf{ 0.22} & -0.20 & -0.25 &\textbf{  0.47} & \textbf{ 0.47} & \textbf{ 0.31} & \textbf{ 0.43} & -0.30 & -0.44 \\ \hline
                \multicolumn{2}{|l|}{$\sigma_{A_m}$}     & -0.06 &  0.01 & -0.11 & -0.15 & \textbf{-0.21} & \textbf{-0.31} & -0.09 & -0.07 & -0.17 & -0.32 & -0.29 & \textbf{-0.50} \\ \hline \hline
    
                \multicolumn{14}{|l|}{Bounding Box \& Mask Criteria} \\
                \parbox[t]{2mm}{\multirow{5}{*}{\rotatebox[origin=c]{90}{Box IoU}}}
                & $iou_{\mathrm{mis}}$                   & \textbf{ 0.29} & \textbf{ 0.34} & \textbf{ 0.25} & \textbf{ 0.29} & \textbf{ 0.27} & \textbf{ 0.30} & \textbf{ 0.45} & \textbf{ 0.56} & \textbf{ 0.38} & \textbf{ 0.49} & \textbf{ 0.39} & \textbf{ 0.49} \\ \cline{2-14}
                & $\mathrm{KL}(p_b|p_m)$                 &  0.07 &  0.19 &  0.10 &  0.16 &  0.06 &  0.07 &  0.08 &  0.10 &  0.15 &  0.28 &  0.12 &  0.19 \\ \cline{2-14}
                & $\mathrm{KL}(p_m|p_b)$                 & -0.09 &  0.18 & -0.06 &  0.13 & -0.07 &  0.07 & -0.12 & -0.04 & -0.07 &  0.22 & -0.08 &  0.17 \\ \cline{2-14}
                & $\mathrm{JS}$                          &  0.14 &  0.14 &  0.12 &  0.13 &  0.12 &  0.13 &  0.15 &  0.25 &  0.19 &  0.25 &  0.20 &  0.26 \\ \cline{2-14}
                & $\mathrm{EMD}$                         & -0.01 & \textbf{ 0.30} &  0.06 & \textbf{ 0.25} &  0.13 &  0.24 &  0.04 &  0.06 &  0.15 & \textbf{ 0.44} &  0.31 &  0.44 \\ \hline
                \parbox[t]{2mm}{\multirow{5}{*}{\rotatebox[origin=c]{90}{Mask IoU}}}
                & $iou_{\mathrm{mis}}$                   & \textbf{ 0.34} & \textbf{ 0.33} & \textbf{ 0.26} & \textbf{ 0.28} & \textbf{ 0.28} & \textbf{ 0.30} & \textbf{ 0.58} & \textbf{ 0.59} & \textbf{ 0.40} & \textbf{ 0.47} & \textbf{ 0.40} & \textbf{ 0.47} \\ \cline{2-14}
                & $\mathrm{KL}(p_b|p_m)$                 &  0.08 & \textbf{ 0.22} &  0.11 &  0.17 &  0.07 &  0.08 &  0.27 &  0.35 &  0.16 &  0.30 &  0.15 &  0.21 \\ \cline{2-14}
                & $\mathrm{KL}(p_m|p_b)$                 & -0.06 & \textbf{ 0.26} & -0.07 &  0.14 & -0.07 &  0.09 &  0.20 &  0.32 & -0.07 &  0.26 & -0.07 &  0.21 \\ \cline{2-14}
                & $\mathrm{JS}$                          &  0.19 &  0.21 & \textbf{ 0.15} &  0.15 &  0.16 &  0.16 &  0.15 &  0.27 & \textbf{ 0.24} &  0.30 &  0.25 &  0.31 \\ \cline{2-14}
                & $\mathrm{EMD}$                         &  0.01 & \textbf{ 0.31} &  0.06 & \textbf{ 0.25} &  0.14 &  0.25 & \textbf{ 0.46} & \textbf{ 0.50} &  0.17 & \textbf{ 0.44} & \textbf{ 0.34} & \textbf{ 0.45} \\ \hline
            \end{tabular}
            \caption{Correlations of \emph{Bounding Box Criteria} and Box IoU, \emph{Instance Mask Criteria} and Mask IoU, and \emph{Bounding Box \& Mask Criteria} with Box IoU and Mask IoU.}\label{tab:sup_correlation}
            \vspace{-10pt}
        \end{table*}
        For the spatial uncertainty, we present three different groups of criteria. The first group covers the bounding box uncertainty (Section~\ref{subsec:box_criteria}), the second the instance mask uncertainty (Section~\ref{subsec:mask_criteria}), and the last group includes criteria resulting from the combination of both uncertainties (Section~\ref{subsec:box_mask_criteria}). We consider the correlation between the IoU and the criteria to evaluate how well the model criteria describe the model performance. The IoU is calculated between GT, the mean bounding box (Box IoU), and the mean instance mask (Mask IoU). Subsequently, the linear correlation is calculated with \emph{Pearson (Pear.)} and the monotonic correlation with \emph{Spearman (Spear.)}. The monotonic correlation is interesting as Box IoU and Mask IoU monotonically increase. Table~\ref{tab:sup_correlation} lists the calculated correlation values. The linear correlation is low, except for a few values. The \emph{Spearman} correlation always delivers better results, but they are moderate. The criteria that stand out the most are $\overline{iou}_b$ for the bounding boxes, $\overline{iou}_m$ for the instance masks, and $iou_{\mathrm{mis}}$ for the combination of the bounding box and mask.

    

\section{Examples}
\begin{figure*}[t]
                \centering
                \begin{subfigure}[h]{1\textwidth}
                    \includegraphics[width=0.24\textwidth]{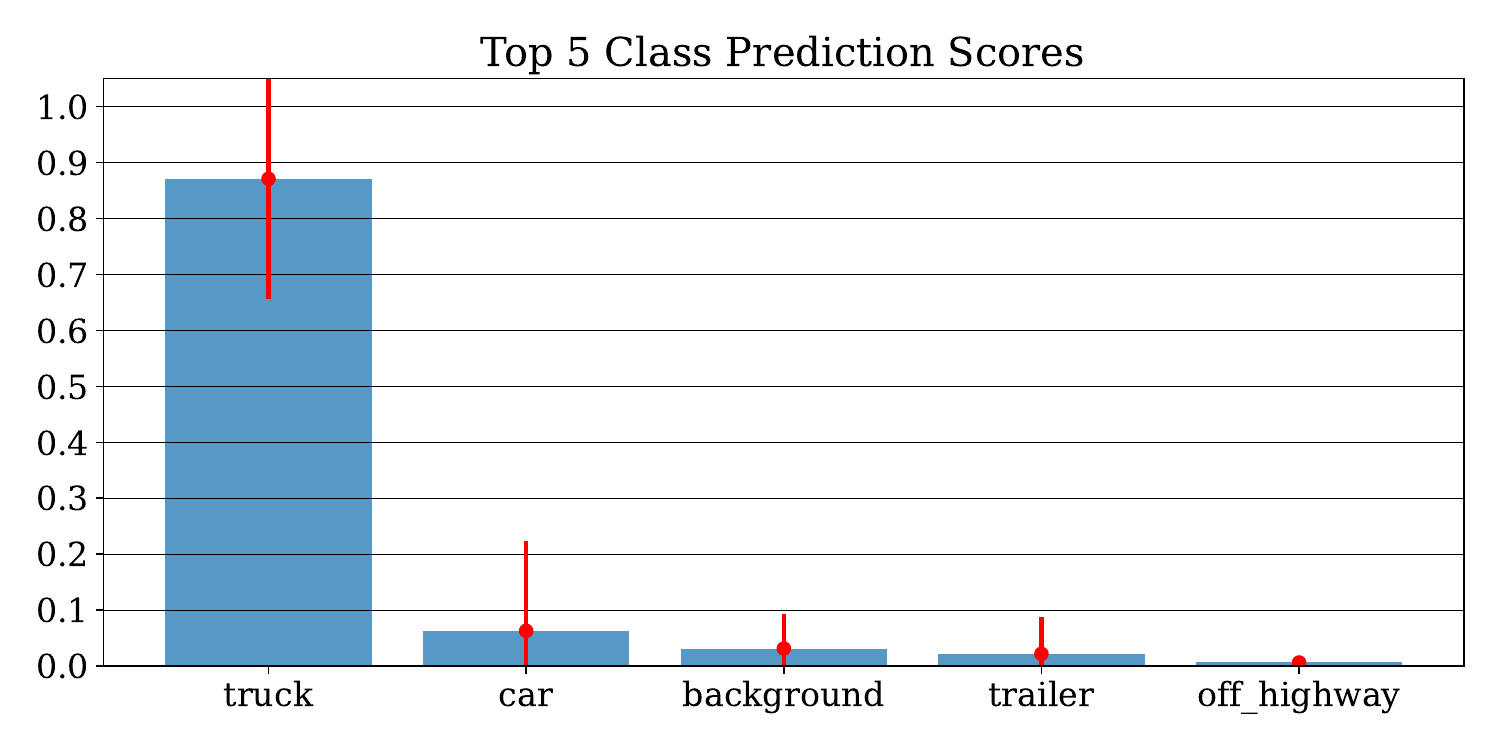}
                    \includegraphics[width=0.245\textwidth]{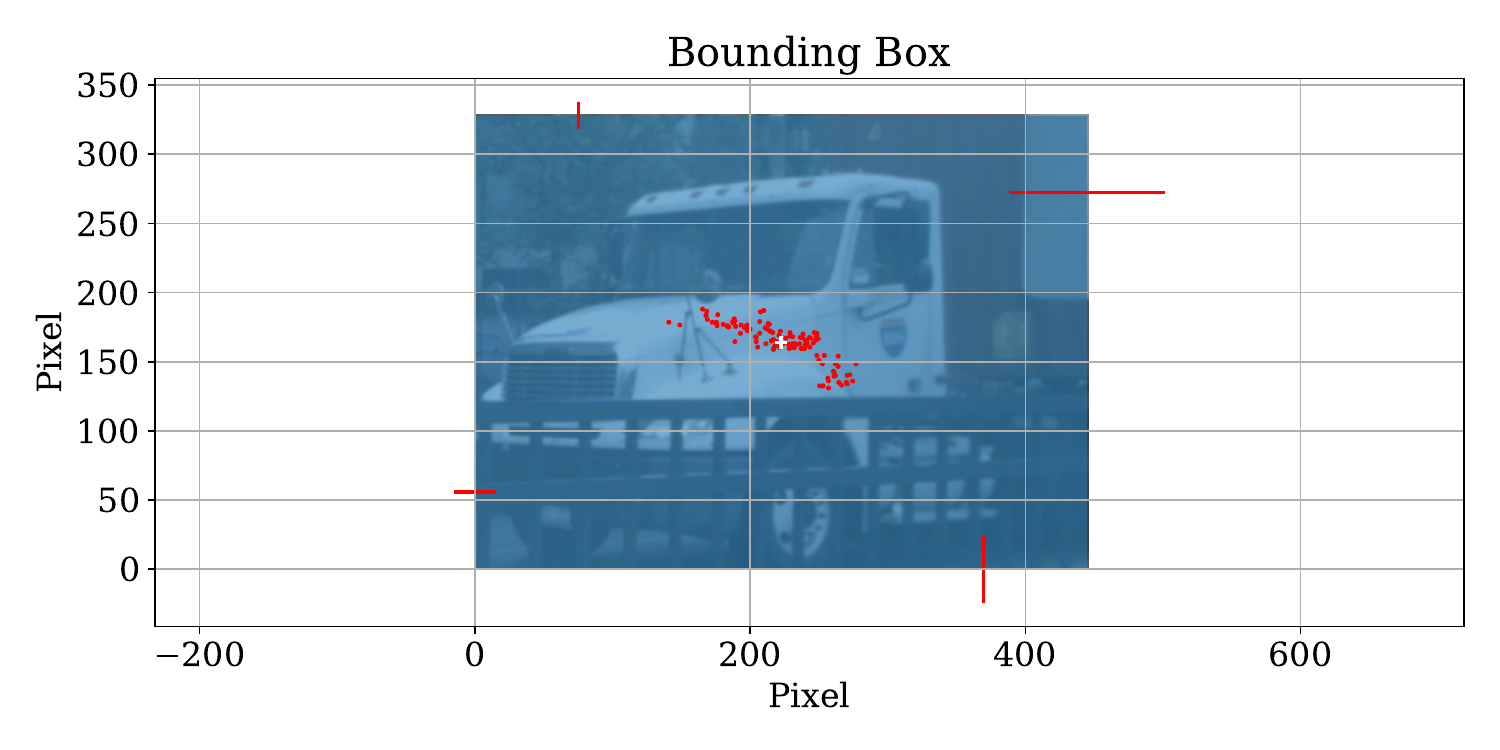}
                    \fbox{\includegraphics[width=0.49\textwidth, trim={0 0 20cm 0},clip]{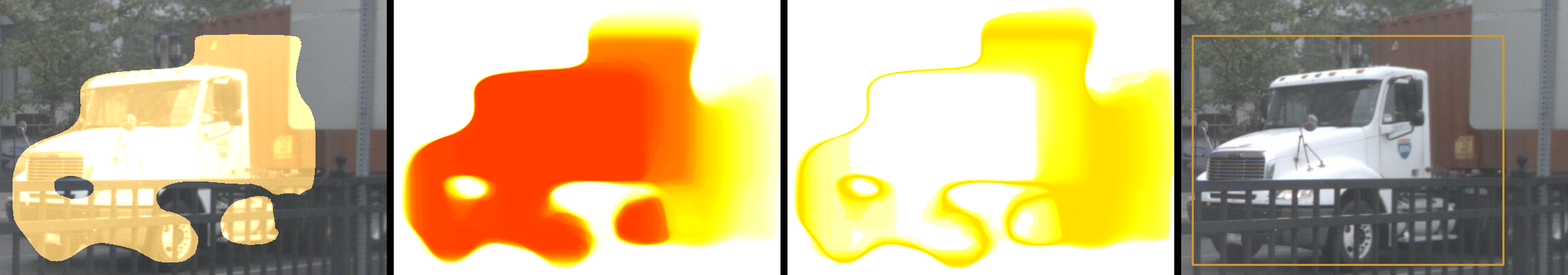}}
                    \caption{True Positive Prediction (TP-P), truck}\label{fig:sup_example_1}
                \end{subfigure}
                
                \vfill
                \begin{subfigure}[h]{1\textwidth}
                    \includegraphics[width=0.24\textwidth]{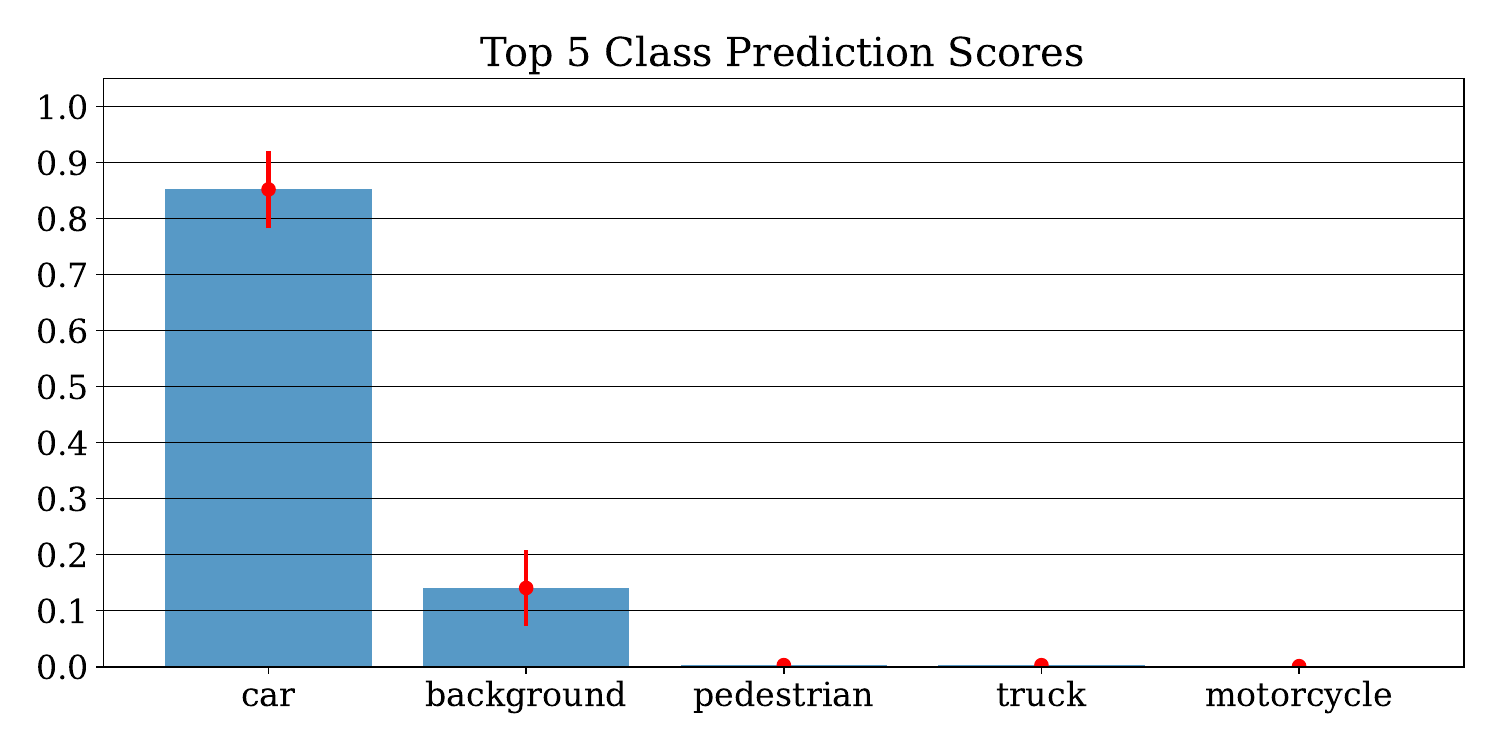}
                    \includegraphics[width=0.245\textwidth]{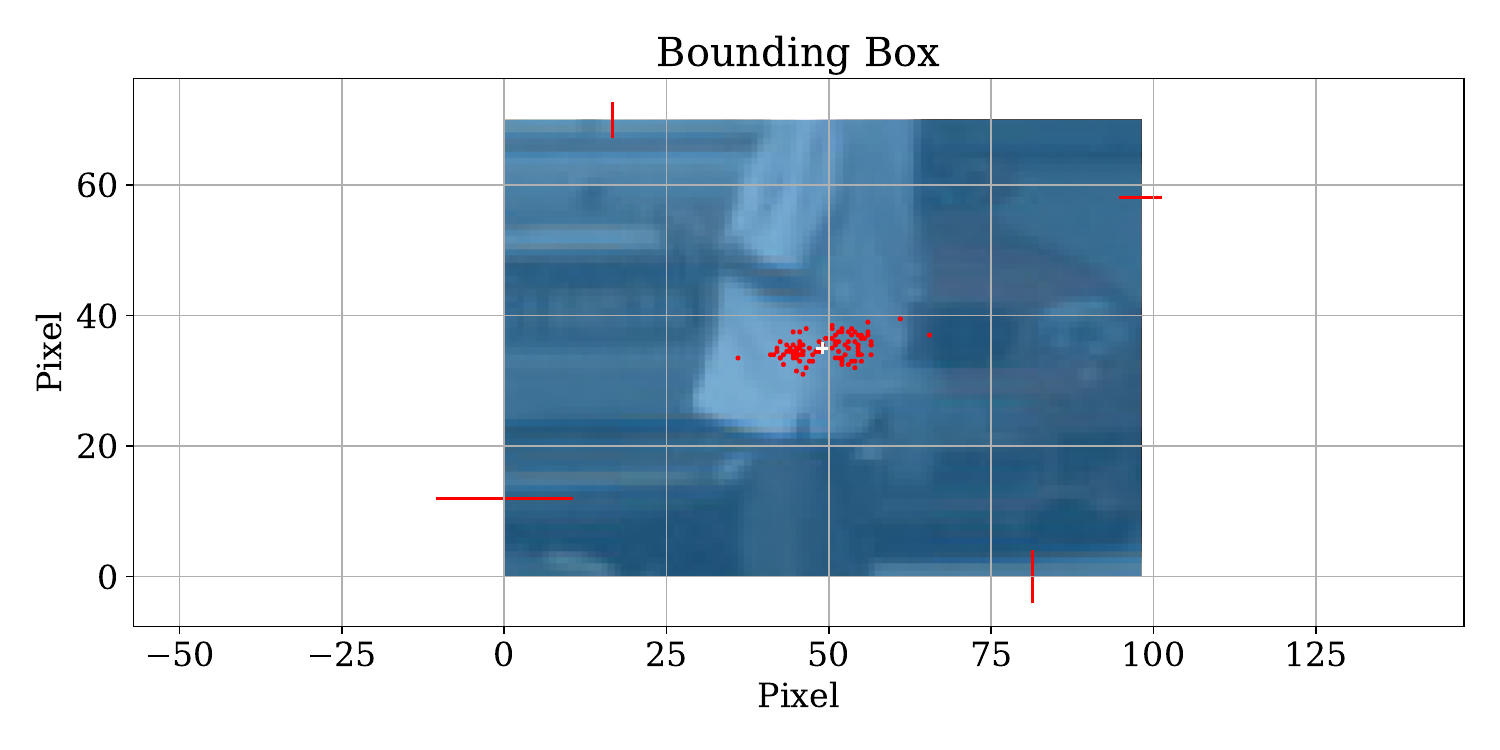}
                    \fbox{\includegraphics[width=0.49\textwidth, trim={0 0 5.5cm 0},clip]{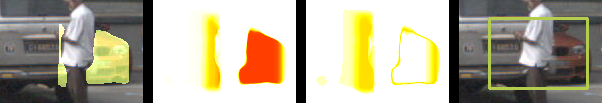}}
                    \caption{True Positive Prediction (TP-P), car}\label{fig:sup_example_3}
                \end{subfigure}
                
                \vfill
                \begin{subfigure}[h]{1\textwidth}
                    \includegraphics[width=0.24\textwidth]{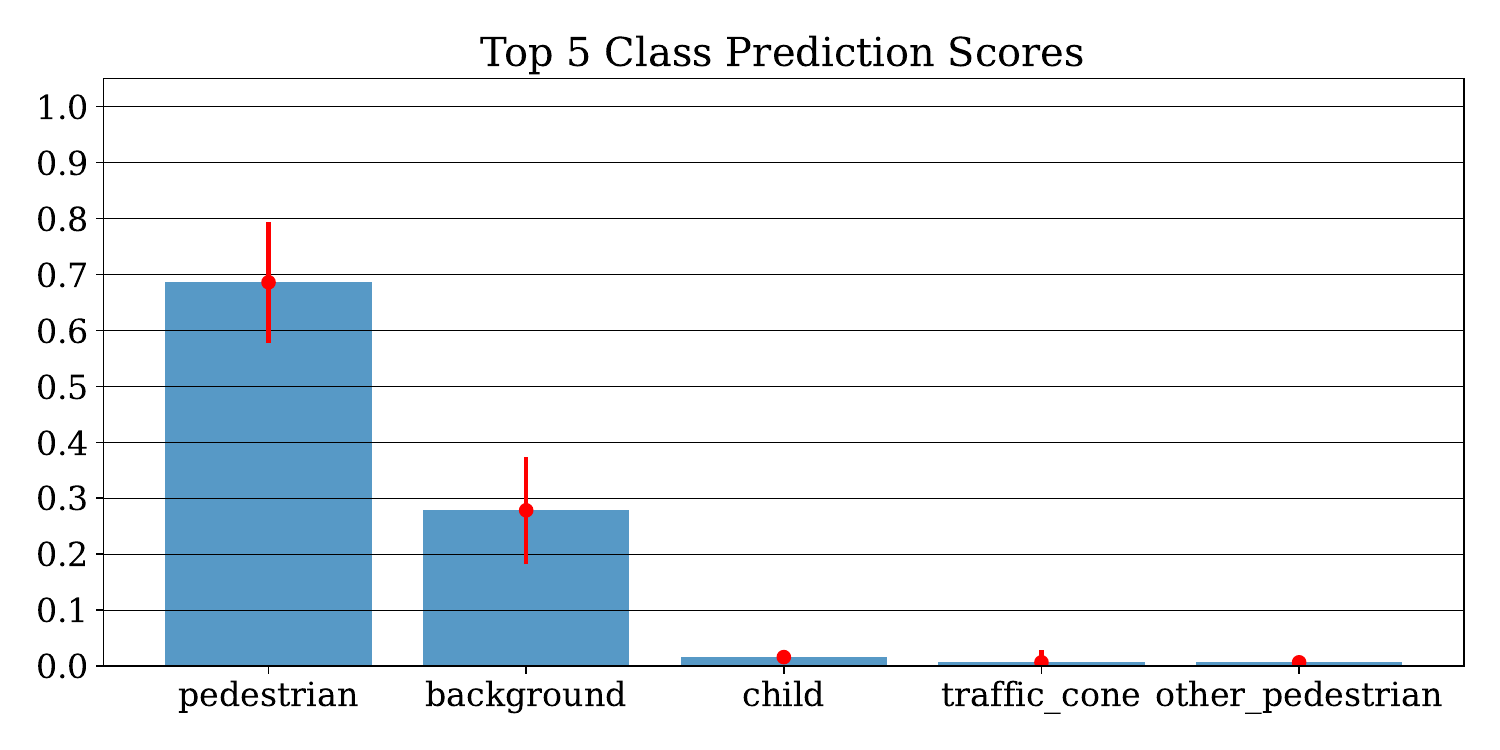}
                    \includegraphics[width=0.245\textwidth]{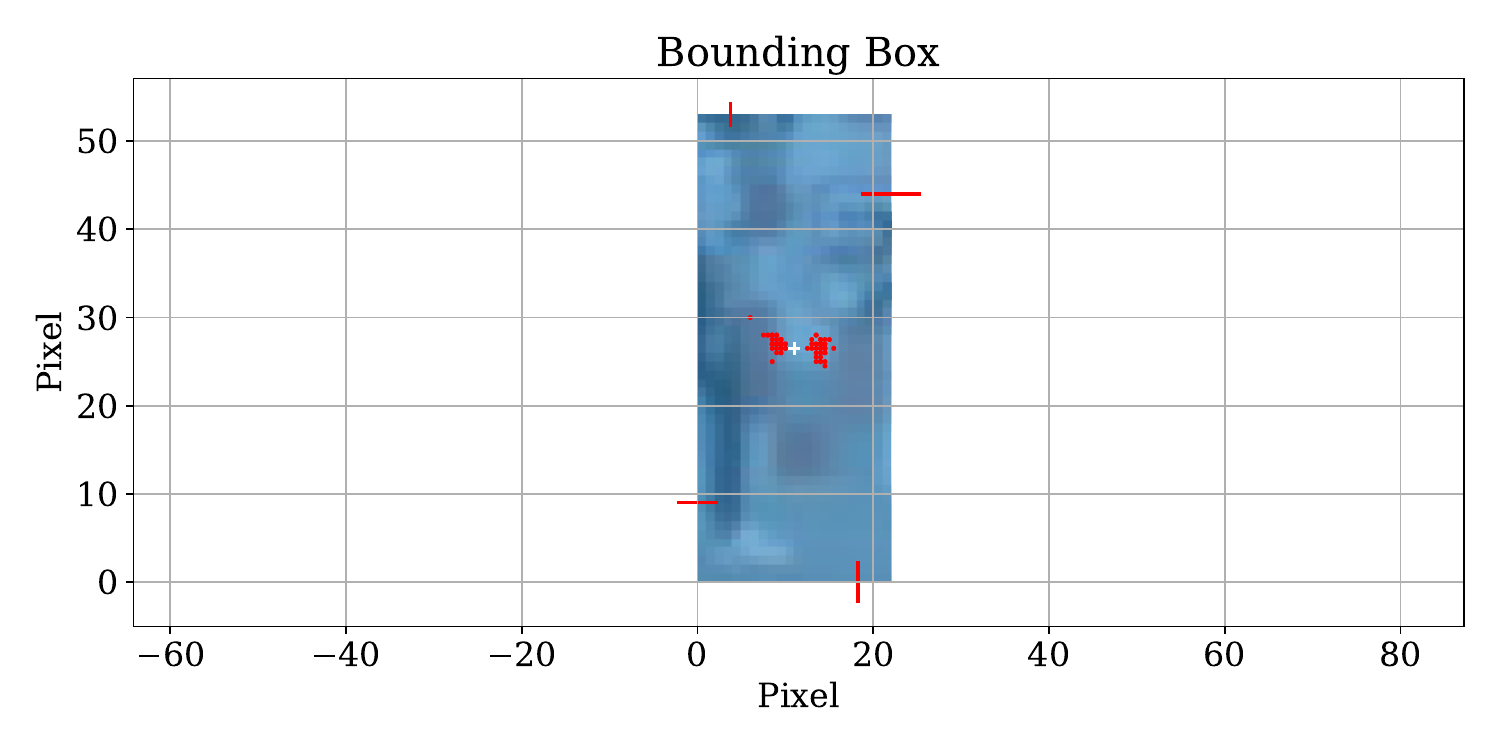}
                    \fbox{\includegraphics[width=0.3\textwidth, trim={0 0 2.1cm 0},clip]{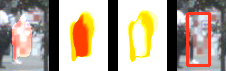}}
                    \caption{Localization Corner Case (L-CC), pedestrian}\label{fig:sup_example_2}
                \end{subfigure}
                
                \vfill
                \begin{subfigure}[h]{1\textwidth}
                    \includegraphics[width=0.24\textwidth]{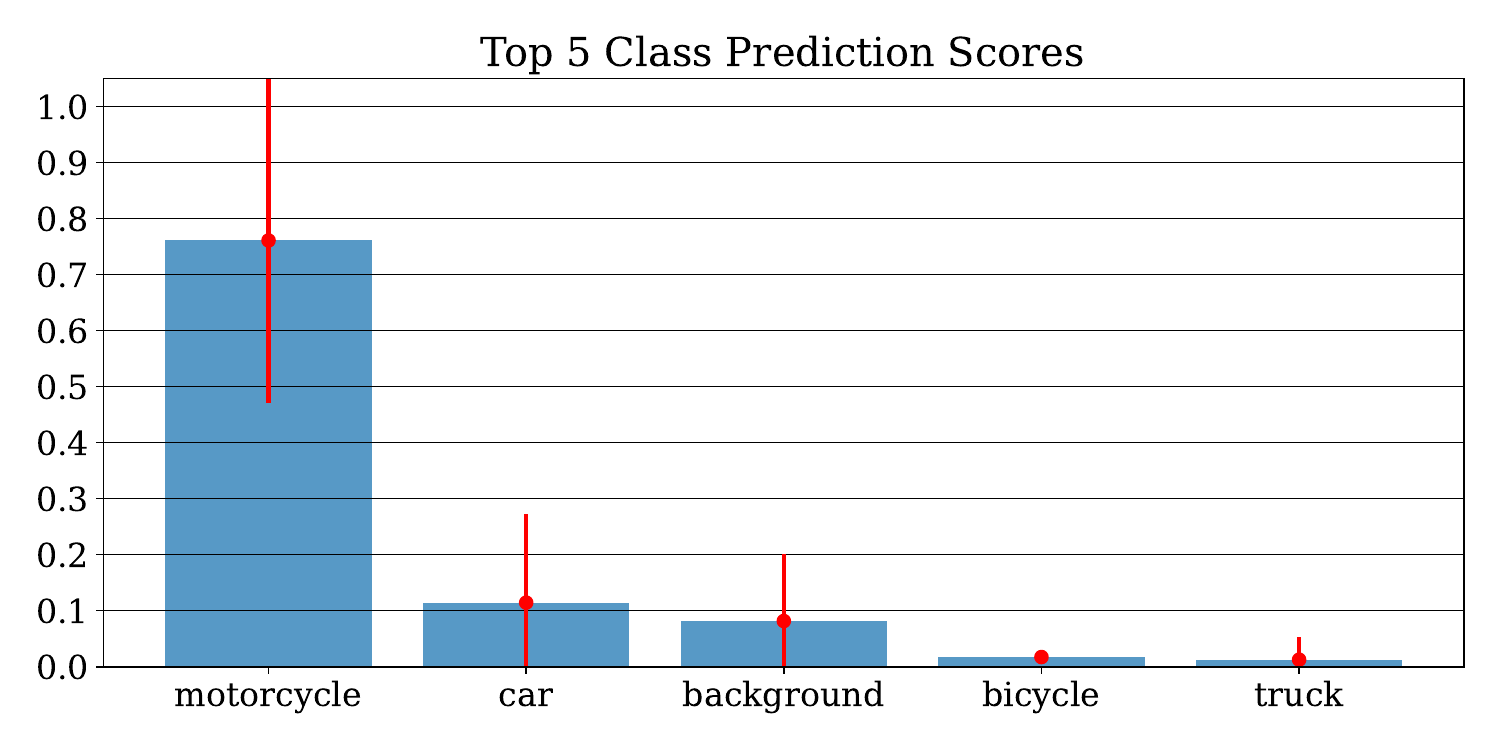}
                    \includegraphics[width=0.245\textwidth]{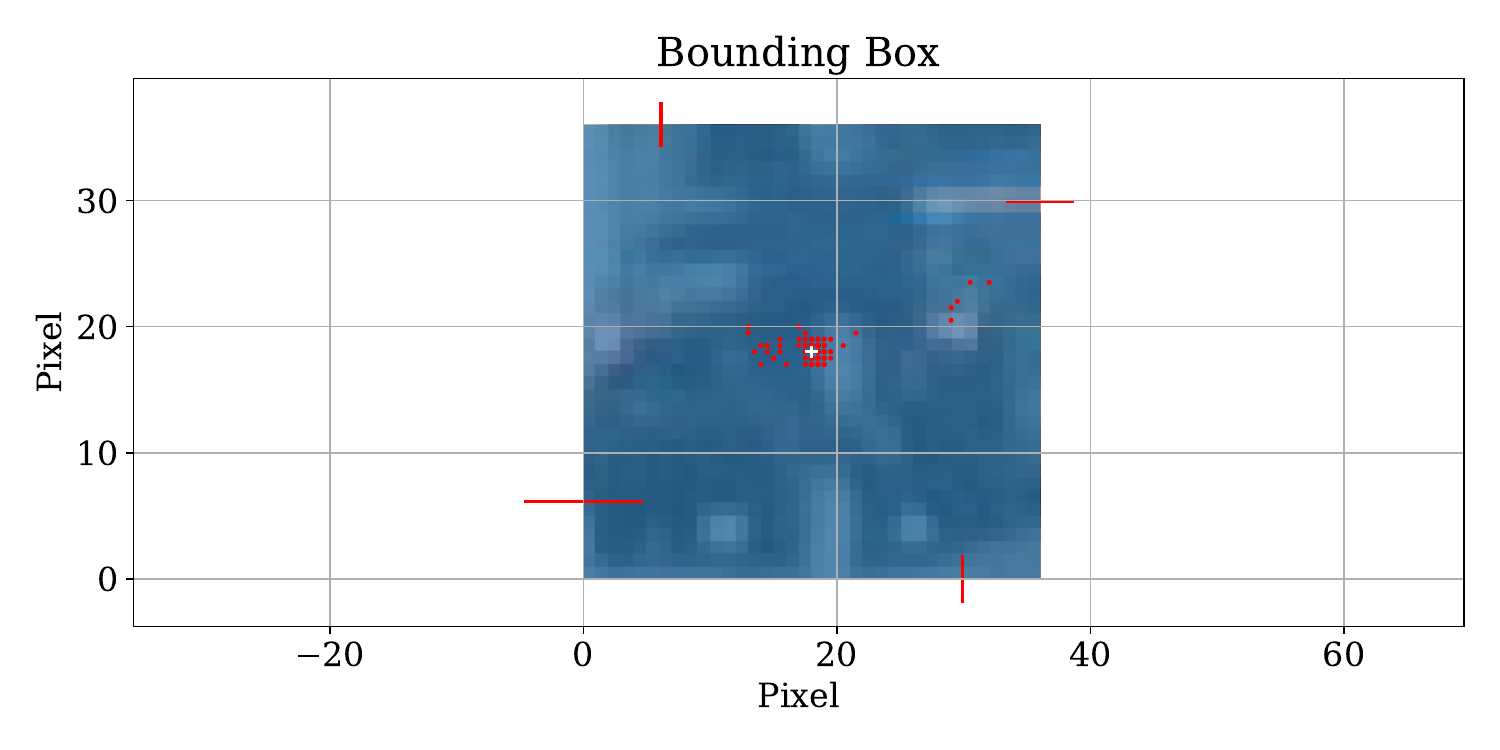}
                    \fbox{\includegraphics[width=0.49\textwidth, trim={0 0 3.5cm 0},clip]{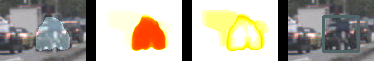}}
                    \caption{Classification Corner Case (C-CC), car}\label{fig:sup_example_4}
                \end{subfigure}
                
                \vfill
                \begin{subfigure}[h]{1\textwidth}
                    \includegraphics[width=0.24\textwidth]{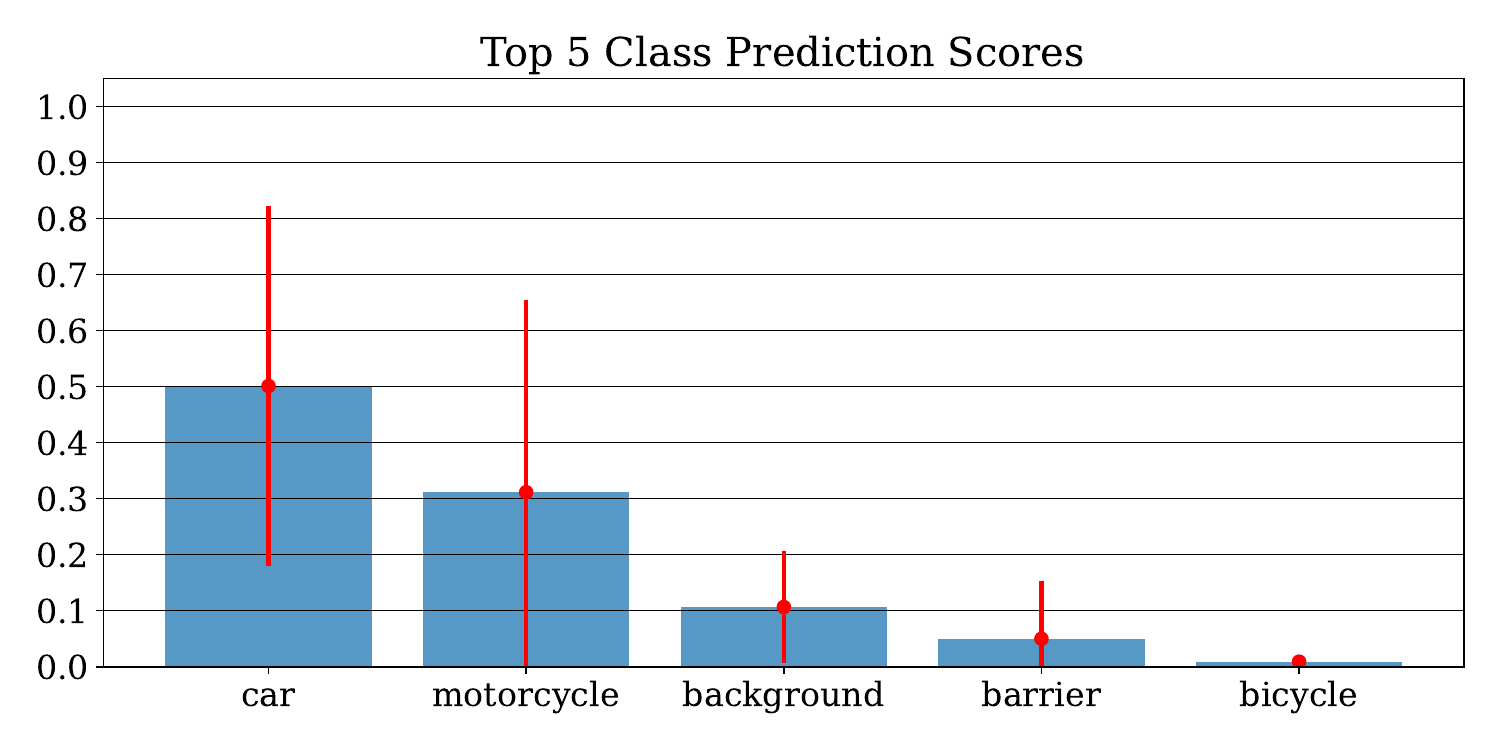}
                    \includegraphics[width=0.245\textwidth]{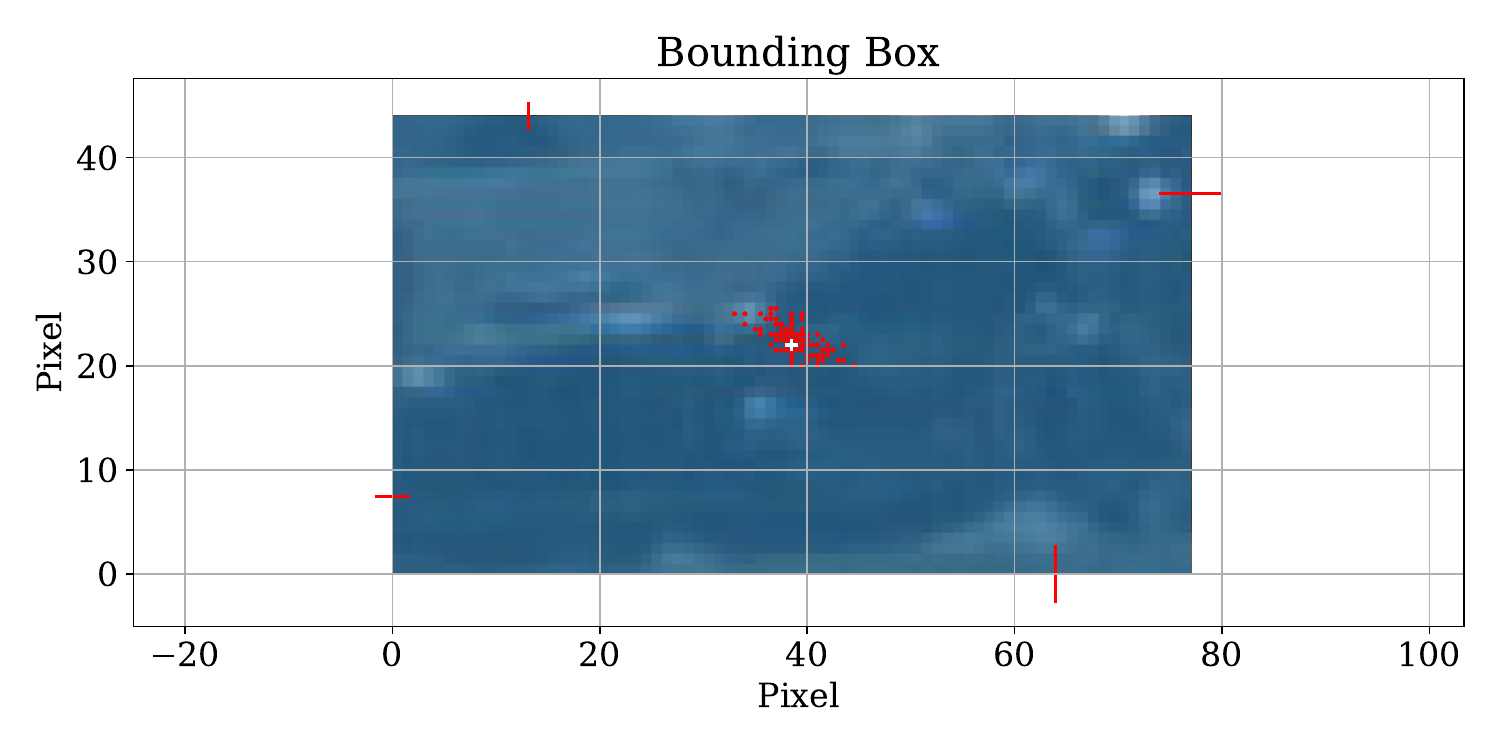}
                    \fbox{\includegraphics[width=0.49\textwidth, trim={0 0 4cm 0},clip]{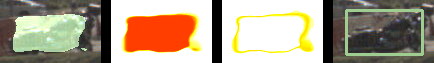}}
                    \caption{Localization \& Classification Corner Case (LC-CC), motorcycle}\label{fig:sup_example_5}
                \end{subfigure}

                \vfill
                \begin{subfigure}[h]{1\textwidth}
                    \includegraphics[width=0.24\textwidth]{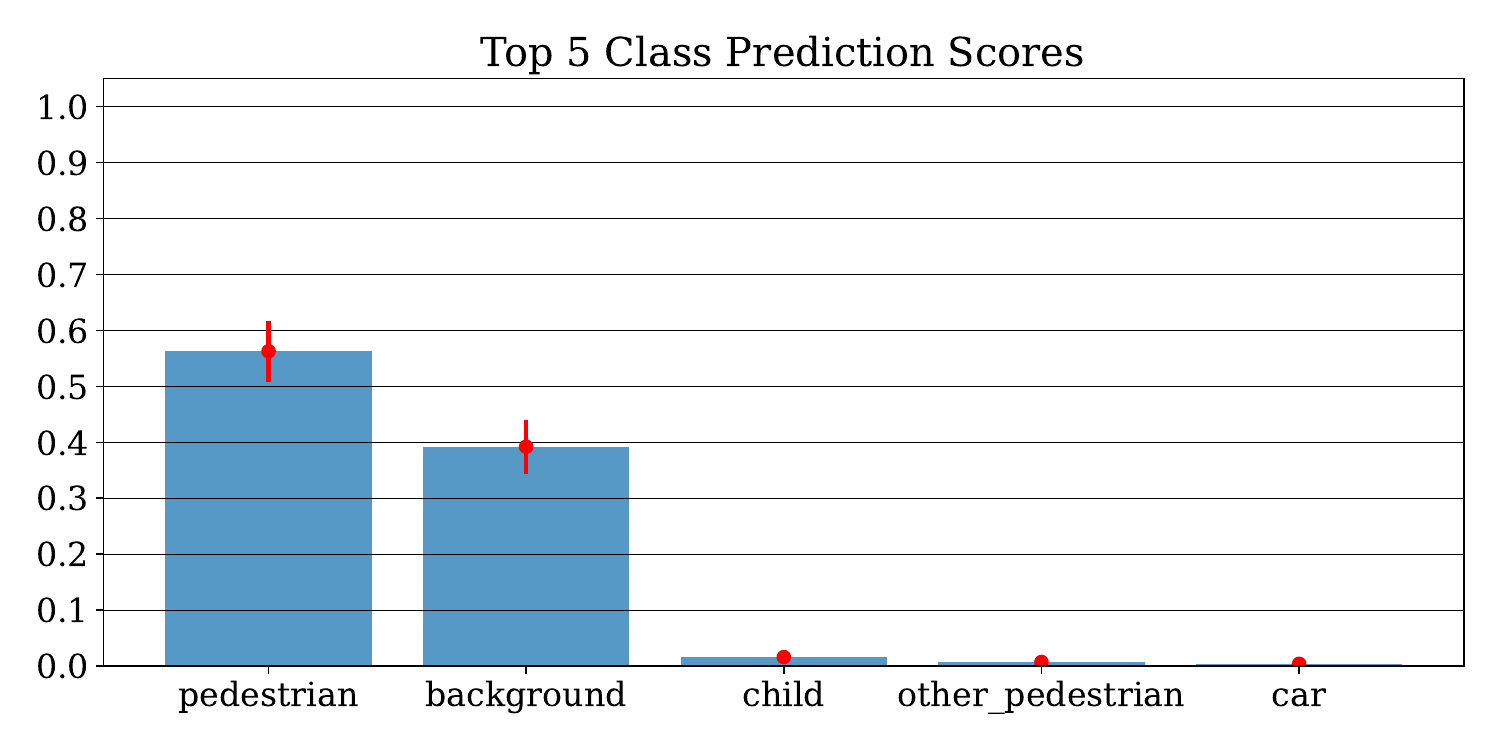}
                    \includegraphics[width=0.24\textwidth]{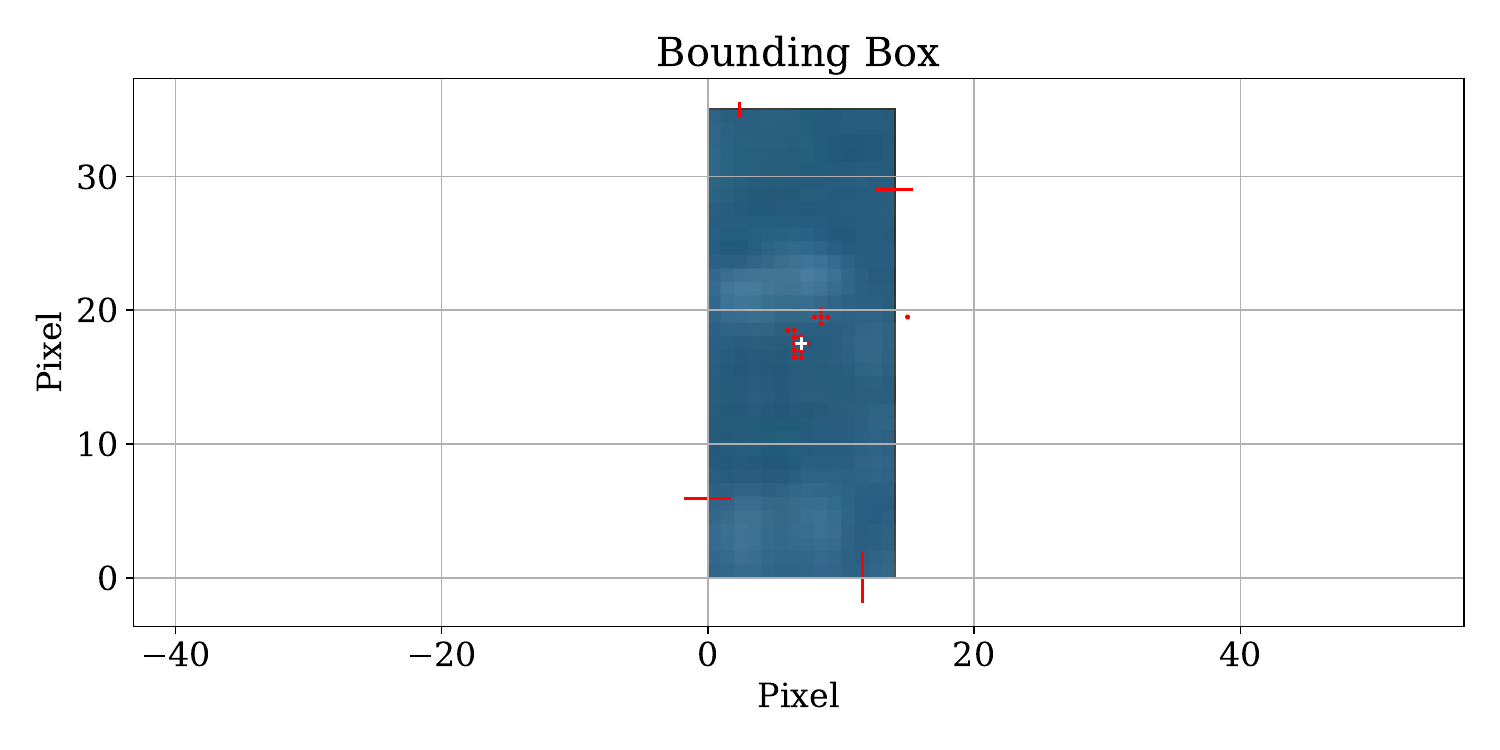}
                    \fbox{\includegraphics[width=0.35\textwidth, trim={0 0 1.8cm 0},clip]{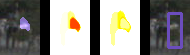}}
                    \caption{False Positive Prediction (FP-P)}\label{fig:sup_example_6}
                \end{subfigure}

                \vfill
                \begin{subfigure}[h]{1\textwidth}
                    \includegraphics[width=0.24\textwidth]{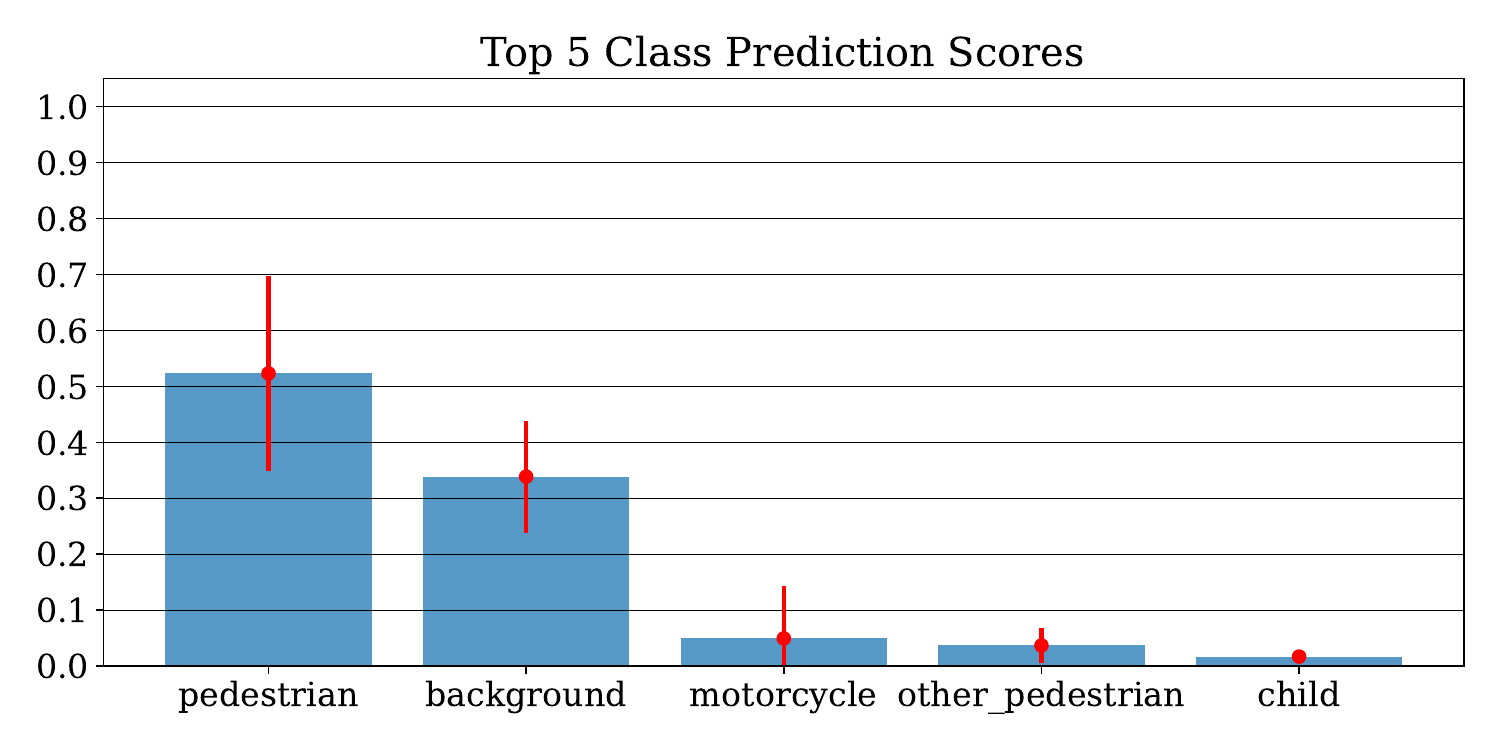}
                    \includegraphics[width=0.24\textwidth]{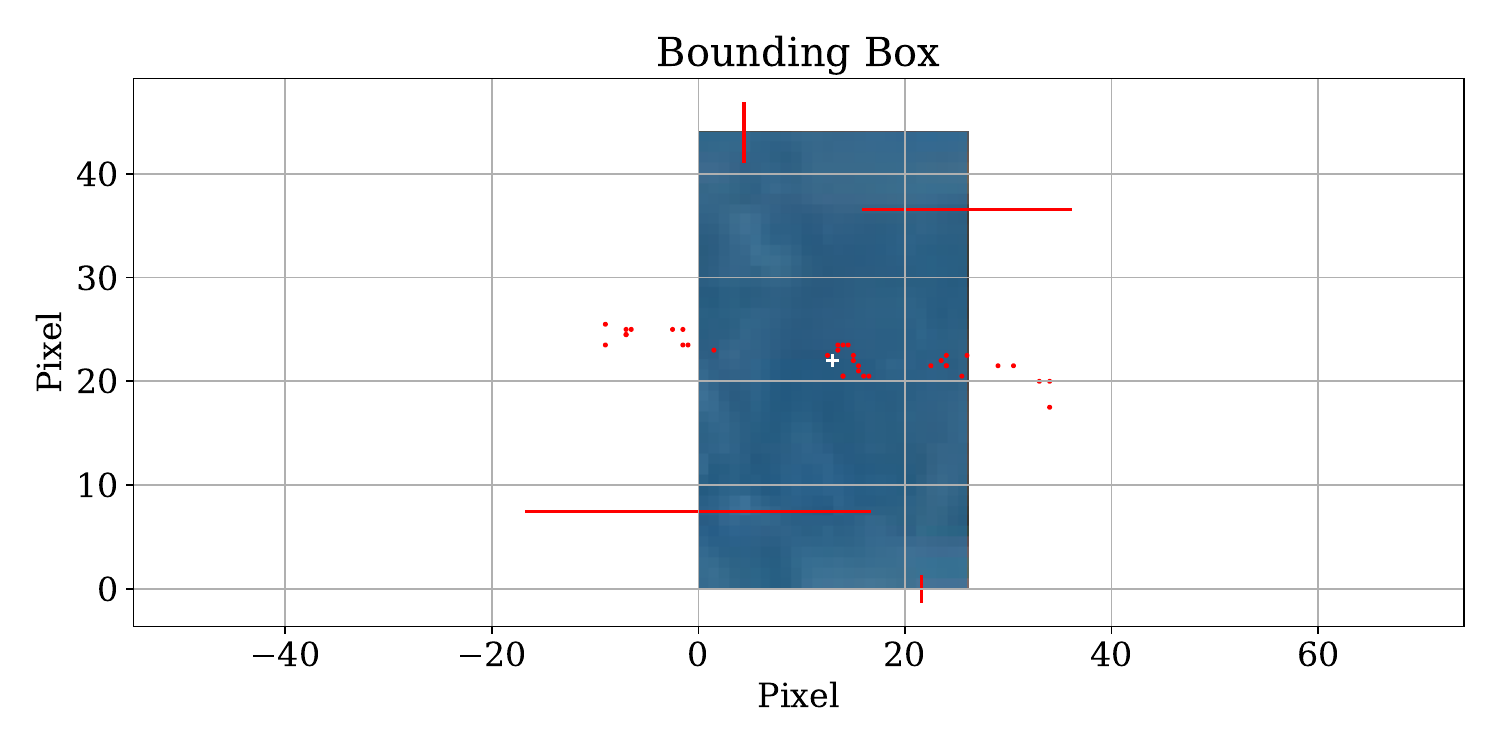}
                    \fbox{\includegraphics[width=0.42\textwidth, trim={0 0 4cm 0},clip]{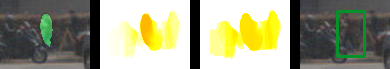}}
                    \caption{False Positive Prediction (FP-P)}\label{fig:sup_example_7}
                \end{subfigure}

                \caption{Uncertainty results of some NuImages examples.}
                \label{fig:sup_examples}
            \end{figure*}

    This section depicts a few examples of object detections (Fig.~\ref{fig:sup_examples}) and the calculated criteria values (Table~\ref{tab:sup_examples}).

\begin{table*}[t]
\centering
\begin{tabular}{|ll|lllllll|} \hline
&          & \multicolumn{7}{c|}{NuImages examples}                         \\ \cline{3-9} 
\multicolumn{2}{|l|}{}                       & 
\multicolumn{1}{l|}{\ref{fig:sup_example_1}} & \multicolumn{1}{l|}{\ref{fig:sup_example_2}} & \multicolumn{1}{l|}{\ref{fig:sup_example_3}} & 
\multicolumn{1}{l|}{\ref{fig:sup_example_4}} & \multicolumn{1}{l|}{\ref{fig:sup_example_5}} & \multicolumn{1}{l|}{\ref{fig:sup_example_6}} & \ref{fig:sup_example_7} \\ \hline
\multicolumn{2}{|l|}{Corner Case Categories} & 
\multicolumn{1}{l|}{TP-P} & \multicolumn{1}{l|}{TP-P} & \multicolumn{1}{l|}{L-CC} & \multicolumn{1}{l|}{C-CC}  & \multicolumn{1}{l|}{LC-CC}  & \multicolumn{1}{l|}{FP-P}  & FP-P  \\ \hline
\multicolumn{1}{|l}{\multirow{2}{*}{GT}}                    & Box IoU  & 
\multicolumn{1}{l|}{0.65 } & \multicolumn{1}{l|}{0.795} & \multicolumn{1}{l|}{0.376} & \multicolumn{1}{l|}{0.752}  & \multicolumn{1}{l|}{0.344}  & \multicolumn{1}{l|}{0.0  }  & 0.0  \\ \cline{2-9} 
\multicolumn{1}{|l}{}                    & Mask IoU & 
\multicolumn{1}{l|}{0.571} & \multicolumn{1}{l|}{0.653} & \multicolumn{1}{l|}{0.222} & \multicolumn{1}{l|}{0.494}  & \multicolumn{1}{l|}{0.136}  & \multicolumn{1}{l|}{0.0  }  & 0.0  \\ \hline \hline
\multicolumn{9}{|l|}{Class Score Criteria}  \\ \hline
\multicolumn{2}{|l|}{$\overline{D}_{c^{k_{max}}}$}                               & 
\multicolumn{1}{l|}{0.871} & \multicolumn{1}{l|}{0.852} & \multicolumn{1}{l|}{0.686} & \multicolumn{1}{l|}{0.761}  & \multicolumn{1}{l|}{0.501}  & \multicolumn{1}{l|}{0.563}  & 0.523  \\ \hline
\multicolumn{2}{|l|}{$\sigma_{c^{k_{max}}}$}                               & 
\multicolumn{1}{l|}{0.214} & \multicolumn{1}{l|}{0.069} & \multicolumn{1}{l|}{0.108} & \multicolumn{1}{l|}{0.29 }  & \multicolumn{1}{l|}{0.322}  & \multicolumn{1}{l|}{0.054}  & 0.174  \\ \hline
\multicolumn{2}{|l|}{$\overline{D}_{c^{k_{2nd}}}$}                               & 
\multicolumn{1}{l|}{0.063} & \multicolumn{1}{l|}{0.14 } & \multicolumn{1}{l|}{0.279} & \multicolumn{1}{l|}{0.114}  & \multicolumn{1}{l|}{0.311}  & \multicolumn{1}{l|}{0.392}  & 0.339  \\ \hline
\multicolumn{2}{|l|}{$\sigma_{c^{k_{2nd}}}$}                               & 
\multicolumn{1}{l|}{0.16 } & \multicolumn{1}{l|}{0.068} & \multicolumn{1}{l|}{0.096} & \multicolumn{1}{l|}{0.158}  & \multicolumn{1}{l|}{0.343}  & \multicolumn{1}{l|}{0.048}  & 0.1    \\ \hline \hline
\multicolumn{9}{|l|}{Bounding Box Criteria}  \\ \hline
\multicolumn{1}{|l}{\multirow{8}{*}{$\sigma_{b}$}}   & $x_1$         & 
\multicolumn{1}{l|}{0.032} & \multicolumn{1}{l|}{0.105} & \multicolumn{1}{l|}{0.098} & \multicolumn{1}{l|}{0.129}  & \multicolumn{1}{l|}{0.02 }  & \multicolumn{1}{l|}{0.123}  & 0.627  \\ \cline{2-9} 
\multicolumn{1}{|l}{}                    & $y_1$         & 
\multicolumn{1}{l|}{0.072} & \multicolumn{1}{l|}{0.055} & \multicolumn{1}{l|}{0.042} & \multicolumn{1}{l|}{0.051}  & \multicolumn{1}{l|}{0.06 }  & \multicolumn{1}{l|}{0.051}  & 0.028  \\ \cline{2-9} 
\multicolumn{1}{|l}{}                    & $x_2$         & 
\multicolumn{1}{l|}{0.125} & \multicolumn{1}{l|}{0.031} & \multicolumn{1}{l|}{0.145} & \multicolumn{1}{l|}{0.072}  & \multicolumn{1}{l|}{0.038}  & \multicolumn{1}{l|}{0.098}  & 0.379  \\ \cline{2-9} 
\multicolumn{1}{|l}{}                    & $y_2$         & 
\multicolumn{1}{l|}{0.027} & \multicolumn{1}{l|}{0.036} & \multicolumn{1}{l|}{0.024} & \multicolumn{1}{l|}{0.047}  & \multicolumn{1}{l|}{0.027}  & \multicolumn{1}{l|}{0.013}  & 0.062  \\ \cline{2-9} 
\multicolumn{1}{|l}{}                    & $c_x$         & 
\multicolumn{1}{l|}{0.068} & \multicolumn{1}{l|}{0.052} & \multicolumn{1}{l|}{0.118} & \multicolumn{1}{l|}{0.077}  & \multicolumn{1}{l|}{0.024}  & \multicolumn{1}{l|}{0.109}  & 0.493  \\ \cline{2-9} 
\multicolumn{1}{|l}{}                    & $c_y$         & 
\multicolumn{1}{l|}{0.042} & \multicolumn{1}{l|}{0.025} & \multicolumn{1}{l|}{0.016} & \multicolumn{1}{l|}{0.03 }  & \multicolumn{1}{l|}{0.027}  & \multicolumn{1}{l|}{0.029}  & 0.038  \\ \cline{2-9} 
\multicolumn{1}{|l}{}                    & $width$         & 
\multicolumn{1}{l|}{0.12 } & \multicolumn{1}{l|}{0.116} & \multicolumn{1}{l|}{0.076} & \multicolumn{1}{l|}{0.14 }  & \multicolumn{1}{l|}{0.038}  & \multicolumn{1}{l|}{0.046}  & 0.316  \\ \cline{2-9} 
\multicolumn{1}{|l}{}                    & $heigth$         & 
\multicolumn{1}{l|}{0.068} & \multicolumn{1}{l|}{0.079} & \multicolumn{1}{l|}{0.059} & \multicolumn{1}{l|}{0.078}  & \multicolumn{1}{l|}{0.075}  & \multicolumn{1}{l|}{0.048}  & 0.058  \\ \hline
\multicolumn{2}{|l|}{$\overline{iou}_b$}                               & 
\multicolumn{1}{l|}{0.822} & \multicolumn{1}{l|}{0.824} & \multicolumn{1}{l|}{0.765} & \multicolumn{1}{l|}{0.849}  & \multicolumn{1}{l|}{0.896}  & \multicolumn{1}{l|}{0.852}  & 0.456  \\ \hline
\multicolumn{2}{|l|}{$\sigma_{iou_b}$}                               & 
\multicolumn{1}{l|}{0.092} & \multicolumn{1}{l|}{0.053} & \multicolumn{1}{l|}{0.06 } & \multicolumn{1}{l|}{0.108}  & \multicolumn{1}{l|}{0.054}  & \multicolumn{1}{l|}{0.113}  & 0.327  \\ \hline \hline
\multicolumn{9}{|l|}{Instance Mask Criteria}  \\ \hline
\multicolumn{1}{|l}{\multirow{4}{*}{$\sigma_{m_{box}}$}}   & $c_x$         & 
\multicolumn{1}{l|}{0.069} & \multicolumn{1}{l|}{0.064} & \multicolumn{1}{l|}{0.119} & \multicolumn{1}{l|}{0.077}  & \multicolumn{1}{l|}{0.024}  & \multicolumn{1}{l|}{0.163}  & 0.678  \\ \cline{2-9} 
\multicolumn{1}{|l}{}                    & $c_y$         & 
\multicolumn{1}{l|}{0.041} & \multicolumn{1}{l|}{0.04 } & \multicolumn{1}{l|}{0.045} & \multicolumn{1}{l|}{0.033}  & \multicolumn{1}{l|}{0.015}  & \multicolumn{1}{l|}{0.127}  & 0.073  \\ \cline{2-9} 
\multicolumn{1}{|l}{}                    & $width$         & 
\multicolumn{1}{l|}{0.123} & \multicolumn{1}{l|}{0.129} & \multicolumn{1}{l|}{0.101} & \multicolumn{1}{l|}{0.126}  & \multicolumn{1}{l|}{0.04 }  & \multicolumn{1}{l|}{0.088}  & 0.308  \\ \cline{2-9} 
\multicolumn{1}{|l}{}                    & $heigth$         & 
\multicolumn{1}{l|}{0.069} & \multicolumn{1}{l|}{0.144} & \multicolumn{1}{l|}{0.056} & \multicolumn{1}{l|}{0.063}  & \multicolumn{1}{l|}{0.035}  & \multicolumn{1}{l|}{0.232}  & 0.078  \\ \hline
\multicolumn{2}{|l|}{$\overline{iou}_m$}                               & 
\multicolumn{1}{l|}{0.835} & \multicolumn{1}{l|}{0.774} & \multicolumn{1}{l|}{0.751} & \multicolumn{1}{l|}{0.896}  & \multicolumn{1}{l|}{0.94 }  & \multicolumn{1}{l|}{0.813}  & 0.343  \\ \hline
\multicolumn{2}{|l|}{$\sigma_{iou_m}$}                               & 
\multicolumn{1}{l|}{0.09 } & \multicolumn{1}{l|}{0.072} & \multicolumn{1}{l|}{0.084} & \multicolumn{1}{l|}{0.134}  & \multicolumn{1}{l|}{0.029}  & \multicolumn{1}{l|}{0.196}  & 0.288  \\ \hline
\multicolumn{2}{|l|}{$\sigma_{A_m}$}                               & 
\multicolumn{1}{l|}{0.159} & \multicolumn{1}{l|}{0.222} & \multicolumn{1}{l|}{0.109} & \multicolumn{1}{l|}{0.078}  & \multicolumn{1}{l|}{0.044}  & \multicolumn{1}{l|}{0.251}  & 0.311  \\ \hline \hline
\multicolumn{9}{|l|}{Bounding Box \& Mask Criteria}  \\ \hline
\multicolumn{2}{|l|}{$iou_{mis}$}                               & 
\multicolumn{1}{l|}{0.952} & \multicolumn{1}{l|}{0.654} & \multicolumn{1}{l|}{0.808} & \multicolumn{1}{l|}{0.918}  & \multicolumn{1}{l|}{0.814}  & \multicolumn{1}{l|}{0.301}  & 0.362  \\ \hline
\multicolumn{2}{|l|}{$KL(p_b|p_m)$}                               & 
\multicolumn{1}{l|}{0.049} & \multicolumn{1}{l|}{0.294} & \multicolumn{1}{l|}{0.254} & \multicolumn{1}{l|}{0.214}  & \multicolumn{1}{l|}{3.057}  & \multicolumn{1}{l|}{0.235}  & 0.162  \\ \hline
\multicolumn{2}{|l|}{$KL(p_m|p_b)$}                               & 
\multicolumn{1}{l|}{0.044} & \multicolumn{1}{l|}{0.697} & \multicolumn{1}{l|}{0.383} & \multicolumn{1}{l|}{0.401}  & \multicolumn{1}{l|}{0.44}  & \multicolumn{1}{l|}{0.959}  & 0.135  \\ \hline
\multicolumn{2}{|l|}{$JS(p_b|p_m)$}                               & 
\multicolumn{1}{l|}{0.107} & \multicolumn{1}{l|}{0.278} & \multicolumn{1}{l|}{0.258} & \multicolumn{1}{l|}{0.236}  & \multicolumn{1}{l|}{0.349}  & \multicolumn{1}{l|}{0.269}  & 0.188  \\ \hline
\multicolumn{2}{|l|}{$EMD(p_b|p_m)$}                               & 
\multicolumn{1}{l|}{0.266} & \multicolumn{1}{l|}{0.424} & \multicolumn{1}{l|}{0.684} & \multicolumn{1}{l|}{0.169}  & \multicolumn{1}{l|}{1.966}  & \multicolumn{1}{l|}{0.42 }  & 0.241  \\ \hline
\end{tabular}
\caption{Corner Case Criteria Results of the NuImages examples form Fig.\ref{fig:sup_examples}.}\label{tab:sup_examples}
\end{table*}

\end{document}